\pdfoutput=1
\documentclass[11pt]{article}

\usepackage[final]{acl}

\usepackage{times}
\usepackage{latexsym}
\usepackage{times}
\usepackage{amsfonts,amssymb}
\usepackage{bbding}
\usepackage{caption}
\usepackage{subcaption}
\usepackage{xcolor}
\usepackage{tikz}

\newcommand*\circled[1]{\tikz[baseline=(char.base)]{
            \node[shape=circle,draw,inner sep=2pt] (char) {#1};}}

\newcommand{\STAB}[1]{\begin{tabular}{@{}c@{}}#1\end{tabular}}

\usepackage[T1]{fontenc}
\usepackage[utf8]{inputenc}

\usepackage{microtype}

\usepackage{microtype}

\usepackage{natbib}  % DO NOT CHANGE THIS AND DO NOT ADD ANY OPTIONS TO IT
\usepackage{caption} % DO NOT CHANGE THIS AND DO NOT ADD ANY OPTIONS TO IT

\usepackage{amsmath}
\usepackage{soul}
\usepackage{graphicx}
\usepackage{tabulary}
\usepackage{multirow}
\usepackage{booktabs}
\usepackage{color}
\usepackage{tcolorbox}
\usepackage{newtxtext}
\usepackage{subcaption}
\usepackage{xspace}
\usepackage{hyperref}
\usepackage{todonotes}
\usepackage{siunitx}
\usepackage{enumitem}
\usepackage{array}
\usepackage{xcolor,colortbl}

\newcolumntype{L}{>{\centering\arraybackslash}m{1.4cm}}
\newcolumntype{H}{>{\centering\arraybackslash}m{4.cm}}
\newcolumntype{X}{>{\arraybackslash}m{5.cm}}
\newcolumntype{B}{>{\arraybackslash}m{15.cm}}
\newcolumntype{F}{>{\arraybackslash}m{8.5cm}}
\newcolumntype{G}{>{\centering\arraybackslash}m{0.5cm}}

\definecolor{Gray}{gray}{0.95}
\newcolumntype{a}{>{\columncolor{Gray}}c}

\newcommand{\cda}{\texttt{Non-LLMDA}}
\newcommand{\gllm}{\texttt{Manual-LLMDA}}
\newcommand{\model}{\texttt{Self-LLMDA}}
\newcommand{\allm}{\model}
\definecolor{LightCyan}{rgb}{0.9254,0.8784,0.8196}
\title{Empowering Large Language Models for Textual Data Augmentation}

  \author{
Yichuan Li$^{1*}$,~~Kaize Ding$^{2*}$,~~Jianling Wang$^{3}$,~~Kyumin Lee$^1$\\ 
$^1$Worcester Polytechnic Institute, $^2$Northwestern University
$^3$Google DeepMind
\\ 
\texttt{\{yli29,kmlee\}@wpi.edu}, \texttt{kaize.ding@northwestern.edu}, \texttt{jianlingw@google.com}
}

\begin{document}
\maketitle 
\def\thefootnote{*}\footnotetext{The first two authors contributed equally to this work. Kaize Ding is the corresponding author.}

\begin{abstract}
With the capabilities of understanding and executing natural language instructions, Large language models (LLMs) can potentially act as a powerful tool for textual data augmentation. However, the quality of augmented data depends heavily on the augmentation instructions provided, and the effectiveness can fluctuate across different downstream tasks. While manually crafting and selecting instructions can offer some improvement, this approach faces scalability and consistency issues in practice due to the diversity of downstream tasks. In this work, we address these limitations by proposing a new solution, which can automatically \textit{generate} a large pool of augmentation instructions and \textit{select} the most suitable task-informed instructions, thereby empowering LLMs to create high-quality augmented data for different downstream tasks.
Empirically, the proposed approach consistently generates augmented data with better quality compared to non-LLM and LLM-based data augmentation methods, leading to the best performance on 26 few-shot learning tasks sourced from a wide range of application domains.

\end{abstract}

\section{Introduction}
Large language models (LLMs) have recently demonstrated their potential in performing data augmentation on text data~\citep{Dai2023ChatAugLC, chung2023increasing, yu2023large, yoo2021gpt3mix}. 
Serving as a semantic-preserving transformation function, LLMs transform original texts based on instructions to create diverse and informative data augmentations. With the augmented data, users can further train a spreadable and affordable model~(e.g. \texttt{OPT}~\citep{zhang2022opt}) to perform specific tasks. 
Unlike traditional heuristic-based methods such as word swapping~\citep{wei2019eda} and model-based methods like back-translation~\citep{fadaee2017data}, LLMs offer great potential to produce more fluent, diverse, and semantically consistent augmentations for text data, owing to their great understanding and generalization capabilities.

\begin{figure}[t!]
    \centering    \includegraphics[width=\linewidth]{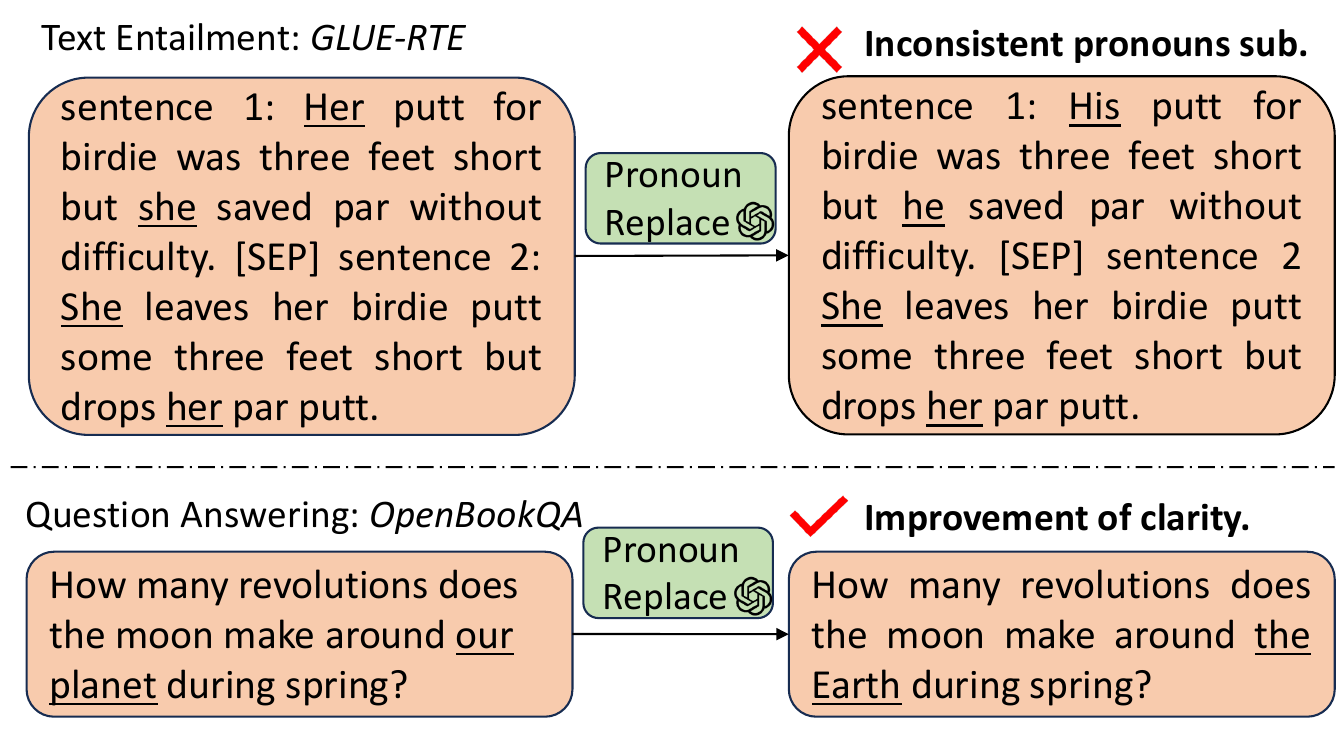}
    
\caption{A simple demo of pronouns replacement augmentation instruction on text entailment task: \textit{GLUE-MRPC}~\citep{wang2019glue} and question answering task: \textit{OpenBookQA}~\citep{mihaylov2018suit}. 
} 
       \label{fig:init_anlysis}
       \vspace{-.5cm}
\end{figure}

Despite the early success of LLMs for textual data augmentation, existing  methods~\cite{Dai2023ChatAugLC} that simply prompt LLMs with human-crafted augmentation instructions (i.e., {\gllm} methods) have the following major bottlenecks: (1) Firstly, their efficacy heavily relies on the quality of the augmentation instructions,  which are manually engineered by domain experts. This manual process is not only domain knowledge-intensive but also prone to inconsistencies, potentially compromising the quality of augmented data. Subtle variations in how these instructions are formulated can significantly influence the outcomes, as demonstrated by recent studies~\citep{ishibashi-etal-2023-evaluating, zhu2023promptbench}; (2) Secondly, usually text augmentation instructions are written in a task-agnostic form for a general purpose, however, the lack of context information on downstream tasks could lead to dramatic performance disparity on different downstream tasks, as shown in \autoref{fig:init_anlysis}. Without considering the specific properties of the target tasks, LLM may generate low-quality augmented data~\citep{ribeiro-etal-2020-beyond, wei2019eda}.

To address the aforementioned challenges, in this paper, we introduce a new framework --{\model} that automates augmentation instruction generation and selection, facilitating LLM to generate task-specific augmented data.
The initial phase of {\model} aims to broaden the span of seed augmentation strategies through the generation of diverse and effective instructions based on LLMs. 
Following this, {\model} employs a scoring model to identify and select the most relevant instructions that are likely to bolster the performance of target models. 
Such a new textual data augmentation approach ensures a balance between the generative breadth of augmentation instructions and targeted precision of task-specific guidance for downstream tasks. 

In our study, we conduct extensive experiments across a large collection of few-shot learning tasks used in previous studies \citep{min-etal-2022-metaicl, ye-etal-2021-crossfit, khashabi-etal-2020-unifiedqa}. This collection includes 26 different types of tasks across hate speech detection, question answering, natural language inference, and phrase detection datasets. 
Our study stands out for its extensive coverage of tasks, setting a new benchmark in the application of LLMs for textual data augmentation when compared to previous work~\citep{Dai2023ChatAugLC, li2023synthetic, chung2023increasing}.
The empirical results demonstrate that the proposed approach {\model} significantly outperforms various baseline methods in generating high-quality augmented textual data. To summarize, our main contributions are as follows:
\begin{itemize}[leftmargin=*,itemsep=0.5pt,topsep=2.5pt]
    \item We introduce a framework {\model}, which automates the generation and selection of task-specific augmentation instructions for LLMs, providing effective data augmentation for text data.
    \item Through a comprehensive set of experiments, we validate the effectiveness of {\model}, demonstrating its superior performance in enhancing data quality and model accuracy over existing text data augmentation methods.
    \item Our in-depth analyses reveal that {\model} can well generalize across various target models and previously unseen augmentation instructions, demonstrating its versatility and potential for broad applicability.
\end{itemize}
\vspace{-0.3cm}

\section{Related Work}
\subsection{Non-LLM Textual Data Augmentation}
Conventional textual data augmentation methods encompass a variety of techniques aimed at enhancing the diversity of textual datasets without relying on large language models (i.e., {\cda} methods).
Those methods range from simple heuristic-based methods to generative model-based methods.
For heuristic-based approaches, such as synonym replacement~\citep{zhang2016characterlevel} and word shuffling, stand out for their computational efficiency and simplicity, making them ideal for large-scale data augmentation with minimal computational demands. 
Another notable example is the Easy Data Augmentation (EDA) technique introduced by \citet{wei2019eda}, which employs token-level perturbations—random insertion, deletion, and swapping—to improve performance across a spectrum of text classification tasks. 

For model-based approaches, researchers have employed seq2seq and language models for data augmentation. 
Back-translation \citep{fadaee2017data} employs translation models to preserve semantic integrity while generating paraphrases \citep{fadaee2017data}.
Conditional masked language models like BERT~\citep{devlin2018bert} and RoBERTa~\cite{liu2019roberta} can also be utilized for data augmentation~\citep{cheng2022semantically, wu2018conditional}. 
By masking words within sentences and subsequently generating replacements, these models introduce linguistic variations. 
Furthermore, other methods \citep{kumar2021data, edwards2023guiding} leverage the capabilities of generative language models like GPT-2 \citep{radford2019language} and BART \citep{lewis2019bart} for data augmentation. These approaches perform conditional generation based on class labels.
Additionally, some studies have explored augmentation in the feature space. Mixup techniques interpolate within word or sentence embeddings \citep{guo2019augmenting}, while others introduce random multiplicative and additive noise to the feature vectors \citep{kurata16b_interspeech}. Despite their utility, these conventional {\cda} methods often come with limitations in readability and contextual consistency.

\begin{figure*}[tbh!]
    \centering
    \includegraphics[width=\linewidth]{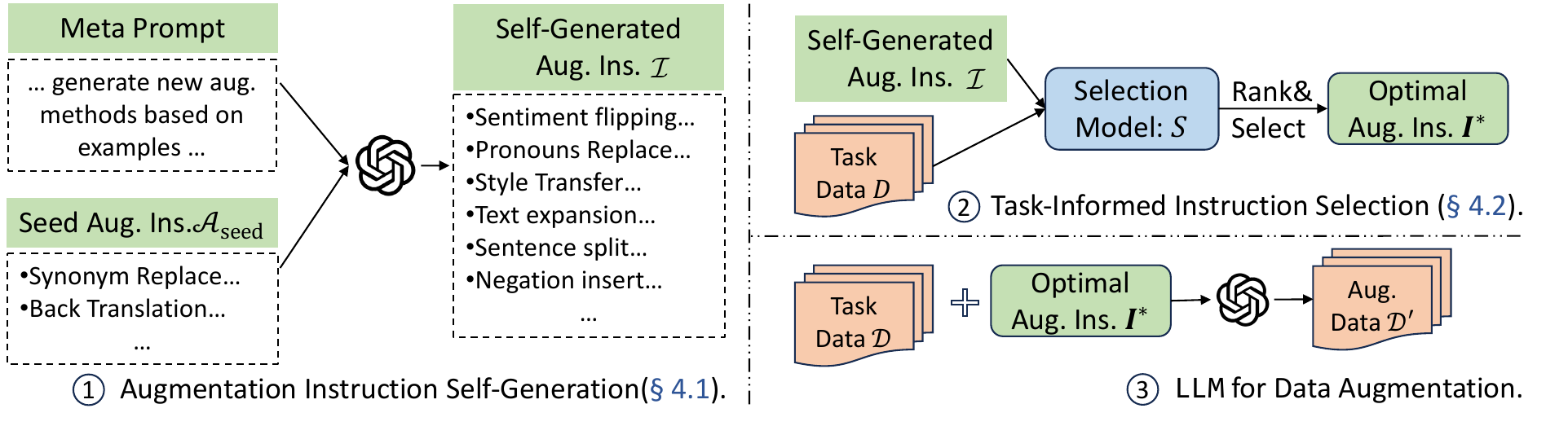}
    \vspace{-0.8cm}
    \caption{The pipeline of {\model}. We first prompt the LLM to generate a diverse set of candidate augmentation instructions (\autoref{sec:ins-gen}). Then we select the instruction (\autoref{sec:selection}) and apply it with the task data to LLM to get augmentations.}
    \label{fig:model-pipe}
    \vspace{-0.45cm}
\end{figure*}

\subsection{LLM-based Textual Data Augmentation}
Recent advancements in LLMs have demonstrated their superiority in generating high quality and contextually relevant augmented data~\citep{brown2020language}.
LLMs are increasingly employed as label-preserving transformation functions, where an original example is transformed or perturbed according to manually crafted instructions~\citep{Dai2023ChatAugLC, yoo2021gpt3mix, piedboeuf-langlais-2023-chatgpt}.
Concurrently, several studies~\citep{chung2023increasing,yu2023large,li2023synthetic,ubani2023zeroshotdataaug,meng2022generating} have explored the generation of conceptually similar yet semantically distinct synthetic examples. 
These methods, however, mostly rely on manual instruction design. In contrast, our work automatically generates label-preserving augmentation instructions by prompting LLMs, thus reducing dependency on manually crafted instructions. Furthermore, we introduce an instruction selection model that chooses appropriate instructions for arbitrary downstream tasks.

\section{Preliminary}
\paragraph{Problem Definition.}
Textual data augmentation involves applying a label-preserving transformation function $T (\cdot)$ to a dataset $\mathcal{D}=\{(\mathbf{x}_i, \mathbf{y}_i)\}_{i=1}^k$, where each example consists of an input text $\mathbf{x}_i$ (a sequence of tokens) and a corresponding label $\mathbf{y}_i$ (also a sequence of tokens). The augmented dataset $\mathcal{D}'$ is generated as follows, ensuring that the output label $\mathbf{y}_i'$ remains unchanged:
\begin{equation}
    \mathbf{x}_i'=T(\mathbf{x}_i), \mathbf{y}_i' = \mathbf{y}_i.
\end{equation}
A target model $F$ is then trained on the union of the original and augmented datasets, $\mathcal{D} \cup \mathcal{D}'$, with the training objective defined as:
\begin{equation}
    \mathcal{L}_{(\hat{\mathbf{x}}_i, \hat{\mathbf{y}}_i)\in{\mathcal{D} \cup \mathcal{D}'}}(F_{\theta}(\hat{\mathbf{x}}_i), \hat{\mathbf{y}}_i).
\end{equation}
Therefore, designing an effective transformation function $T(\cdot)$ that produces high-quality augmented data $\mathcal{D}'$ is crucial for improving the downstream performance of model $F_\theta$.
\paragraph{\gllm.}
For {\gllm} methods, the transformation function $T(\cdot)$ is realized through a combination of an LLM and a manual-crafted instruction $\mathbf{I}_{\text{man}}$~(e.g., paraphrasing). The LLM is prompted to generate semantic-preserving transformations of the input text $\mathbf{x}_i$ for the augmented dataset $\mathcal{D}'$:
\begin{equation}
    \mathbf{x}_i'=\text{LLM}(\mathbf{I}_{\text{man}}, \mathbf{x}_i), \mathbf{y}_i' = \mathbf{y}_i
\end{equation}

\section{Proposed Approach -- \model}
To reduce the human efforts in designing augmentation instructions and selecting a task-specific instruction for a given task, we propose {\model} depicted in \autoref{fig:model-pipe}.  
The process begins with the LLM generating a diverse set of potential instructions $\mathcal{I}=\{\mathbf{I}_j\}_{j=0}^n$ from a given set of seed instructions $\mathcal{I}_{\text{seed}}=\{\mathbf{I}_{\text{man}}\}$:
\begin{equation}
    \mathcal{I} = \text{LLM}(\mathcal{I}_{\text{seed}}).
\end{equation}
A selection model $S$ then scores these generated instructions against the dataset $\mathcal{D}$ to identify the most suitable instruction $\mathbf{I}^*$:
\begin{equation}
    \mathbf{I}^* = S(\mathcal{I}, \mathcal{D}).
\end{equation}
Based on the selected instruction $\mathbf{I}^*$, the LLM performs data augmentation on $\mathcal{D}$, producing an enhanced augmented dataset $\mathcal{D}'$ for training the target model more effectively.

\subsection{Augmentation Instruction Self-Generation }
\label{sec:ins-gen}
Inspired by the self-instruct methodology \cite{wang2022self}, this phase generates augmentation instructions from a seed set of 13 human-crafted instructions. 
These seed instructions act as exemplars, guiding the LLMs toward the creation of novel and diverse instructions that maintain the semantic integrity of the input text. 
To generate a broad and diverse set of augmentation instructions without the bias introduced by a few task examples, we exclude the task-specific data from the instruction generation. 
This will leverage the zero-shot learning capabilities of LLMs to produce a wide array of potential augmentation instructions. 
We use the following prompt to encourage LLMs to explore various augmentation techniques:
\begin{tcolorbox}[boxsep=0mm,left=2.5mm,right=2.5mm,colframe=black!55,colback=black!5]
``\emph{Come up with a series of textual data augmentation methods and you need to generate more \textbf{diverse} data augmentation method that can \textbf{keep the semantic meaning} of the input sentence.
$\{\mathbf{I}_{\text{seed}}\}$} ''
\end{tcolorbox}
Through iterative cycles of generation and refinement, we filter out instructions that are too similar to existing ones based on ROUGE-L~\citep{lin-2004-rouge}. 
The unique generated instructions from each iteration are then incorporated back into the seed instruction pool, enriching the seed instructions for subsequent generation rounds.
This process is repeated until we reach a collection of 100 augmentation instructions. To ensure diversity and eliminate redundancy, we further refine this set by removing duplicates based on their method names. 
This filtration results in a final set of 51 unique augmentation instructions.

\subsection{Task-Informed Instruction Selection}
\label{sec:selection}
\begin{figure}
    \centering
    \includegraphics[width=.9\linewidth]{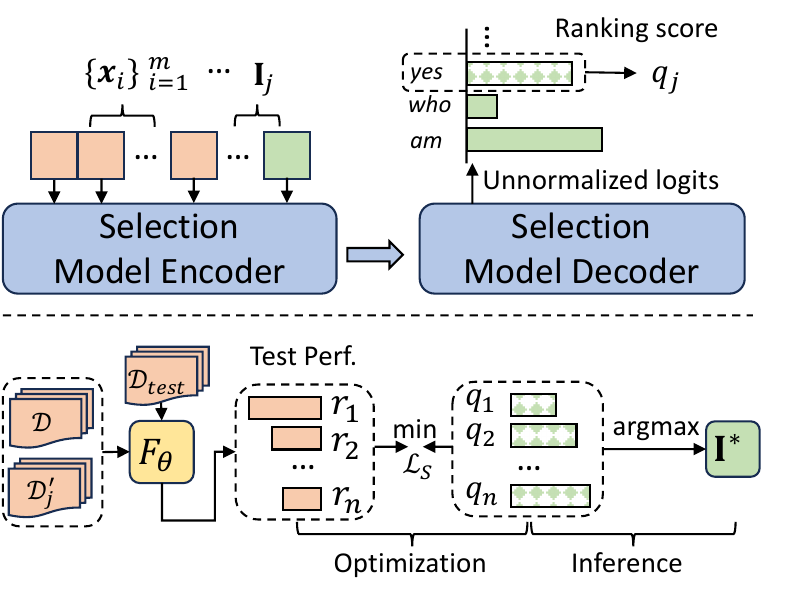}
    \vspace{-.3cm}
    \caption{Illustration of the Instruction selection scoring model. $F_{\theta}$ is the target model. 
    }
    \label{fig:selection}
    \vspace{-.5cm}
\end{figure}

Recognizing that augmentation instructions may not be universally applicable across different tasks, we implement a selection mechanism, tailored to the specific requirements of each task and its corresponding target model. 
This process involves a scoring model $S$ to evaluate the suitability of each instruction for the task at hand. The scoring model $S$,  as shown in \autoref{fig:selection}, outputs a ranking score  $q_j$ indicating the instruction's effectiveness based on the pair of instruction and task dataset. 
Based on the notable instruction-following capabilities of FLAN-T5 \cite{chung2022scaling, raffel2023exploring},  we choose FLAN-T5-Large~\cite{chung2022scaling, raffel2023exploring} as the backbone of our scoring model.
The input for scoring model $S$ is:
\begin{tcolorbox}[boxsep=0mm,left=2.5mm,right=2.5mm,colframe=black!55,colback=black!5]
``\textit{Given the dataset for \textbf{task} ${\mathcal{T}}$ and the instruction data, determine if this is a suitable instruction to address the task for \textbf{model} $F$. \textbf{Task Dataset:} ${{\{\mathbf{x}_i\}}_{i=0}^{m}}$
\textbf{Instruction:} ${\mathbf{I}_j}$. Is this instruction appropriate?}''
\end{tcolorbox}
\noindent where $\mathcal{T}$ is the task name (e.g. GLUE-RTE), $F$ is the target model name (e.g. {OPT-125m}). Since most of the tasks did not have a task description and manually designing the task description is time consuming, we utilize the few-shot examples $\{\mathbf{x}_i\}_{i=0}^m$ from the dataset as the task description. 
Here, we calculate  $q_j$ by assessing the logit value of the ``yes'' token of the last position of input from FLAN-T5-Large, as shown in \autoref{fig:selection}.
Next, we will introduce the optimization and inference procedure of the scoring model $S$ respectively.

\paragraph{Model Optimization.} 
The instruction selection model is trained to prioritize generated augmentation instructions based on their impact on downstream task performance. 
Its goal is to assign the highest scores to instructions that lead to the most effective data augmentation.
To enhance scalability and computational efficiency, our model optimizes the selection process for a given task  $\mathcal{D}$ by sampling a subset of augmentation instructions $\{\mathbf{I}_j\}_{j=0}^{n}$ (where $n>1$) from the pool of candidates.
The model then computes scores $\{q_j\}_{j=0}^n$, representing the relative effectiveness of each instruction. 
The optimization objective is formulated as a cross-entropy loss, designed to accurately distinguish between the effectiveness of these instructions $\{\mathbf{I}_j\}_{j=0}^{n}$. The loss function is given by:
\begin{equation}
    \mathcal{L}_{S} =  -\sum_{j=0}^n \text{is\_max}(r_j) \log \sigma(q_j)
    \label{eq:optimize_s}
\end{equation}
Here, \text{is\_max} serves as a binary indicator function that identifies the instruction yielding the maximum effectiveness, and  $\sigma$ is \texttt{softmax} that normalizes the $q_j$ (a probability generating ``yes'' token interpreted as a ranking score associated with $j$th instruction) over the sampled augmentation instructions, $r_j$ is the downstream task performance of a trained target model on augmented data $\mathcal{D}_j'\cup\mathcal{D}$.

\paragraph{Model Inference.} 
When encountering a new task, the selection model $S$ evaluates all potential instructions
to determine the most suitable one $\mathbf{I}^*$, denoted by the highest score: 
\begin{equation}
   \mathbf{I}^* =  \mathbf{I}_{\text{argmax}(\{q_j\}_{j=0}^{|\mathcal{I}|})}   
\end{equation}
This optimal instruction, $\mathbf{I}^*$, is then employed to prompt the LLMs to generate augmented data. 
This selection mechanism ensure the use of the most effective instruction for enhancing data utility across diverse NLP tasks.

\section{Experiment}

\subsection{Experimental Setup}

\paragraph{Evaluation Datasets.}
In this study, we select 26 few-shot learning tasks spanning a wide range of NLP challenges, sourced from CrossFit \citep{ye-etal-2021-crossfit}, UnifiedQA \citep{khashabi-etal-2020-unifiedqa}, and MetaICL \citep{min-etal-2022-metaicl}. 
These datasets were chosen for their diversity, encompassing both classification tasks (Class)—such as natural language inference, paraphrase detection, and hate speech identification—and non-classification (Non-Class) tasks, notably question answering, to ensure a broad evaluation spectrum. 
The selection of tasks is significantly larger and more diverse than that in other relevant works \citep{Dai2023ChatAugLC,chung2023increasing}.

To investigate the generalization ability of {\model}, we split the 26 tasks into training and test tasks as for form ``train$\rightarrow$ test''. We train the augmentation instruction selection methods on training tasks and evaluated it on test tasks. 
The task split involves four settings: Class $\rightarrow$ Class, Class $\rightarrow$ Non-Class, Non-Class$\rightarrow$ Class, and Random $\rightarrow$ Random, where ``Random'' represents a mixture of randomly selected tasks~\footnote{Details of training and testing tasks split is in \autoref{tab:tasks_all}. }. 
This design allows us to investigate the performance of selection models when applied across similar and disparate task types, providing insights into their generalizability and effectiveness.

\paragraph{Evaluation Metrics.}
To handle all types of tasks simultaneously, we unify all downstream tasks, including classification and non-classification tasks, using a text-to-text approach~\citep{raffel2023exploring}.  For each task, we feed the input text to the target model $F_\theta$ and train it to generate the corresponding target text.
We choose \texttt{OPT}~\citep{zhang2022opt} from three different sizes (e.g. 125m, 350m and 1.3b\footnote{
Due to GPU memory constraints, the training mini batch size for the 1.3B model is set to 2,  while the batch sizes for the 125M and 350M models are set to 8. This difference in batch sizes may cause the 1.3B model to achieve worse performance compared to the 125M and 350M models.
 }) as our target models $F_\theta$. 
During training\footnote{Detailed hyparameter setting is in \autoref{sec:detail_exp_setting}.}, $F_\theta$ takes the training example $\mathbf{x}_i$ as the input, and is optimised to generate $\mathbf{y}_i$ using the negative likelihood objective function:
\begin{equation}
    \mathcal{L}_{F_{\theta}}(\mathbf{y}_i) = -\sum_{t=1}^{|\mathbf{y}_i|}\log P_{F_{\theta}}(y_i^t|\mathbf{x}_i, \mathbf{y}_i^{<t})
\end{equation}
During inference time, given the test input $\mathbf{x}_{test}$ as well as a set of candidates $\mathcal{C}$, which is either a set of labels (in classification tasks) or answer options (in non-classification tasks), the $F_\theta$ computes the conditional probability of each label $\mathbf{c} \in \mathcal{C}$, where $\mathbf{c}$ is a sequence of tokens. The label with the maximum conditional probability is returned as a prediction:
\begin{equation}
    \text{argmax}_{\mathbf{c} \in \mathcal{C}}\left( \sum_{t=1}^{|\mathbf{c}|}\log P_{F_{\theta}}(c^t|\mathbf{x}_{\text{text}}, \mathbf{c}^{<t})\right)
\end{equation}

Specifically, we use \textit{macro-F1} for classification tasks, and \textit{accuracy} for non-classification tasks in our experiment. 
The overall performance is then quantified by computing the macro-average of these scores across all tasks, encapsulating both \textit{accuracy} and \textit{macro-F1} metrics. 
To ensure robustness and reduce sampling bias, each experiment under each splitting setting is replicated with five different random seeds. 
For each few-shot task, we adopt a uniform approach by randomly selecting $k=16$ training examples. Following \citep{min-etal-2022-metaicl}, we did not make perfect label balance between $k$ training examples. 
\begin{table*}[tbh!]
    \centering
    \small
    \scalebox{.9}{
    \begin{tabular}{cl aaa ccc aaa ccc}
    \toprule
         & {TextulDA} & \multicolumn{3}{c}{Class $\rightarrow$ Class} & \multicolumn{3}{c}{Class $\rightarrow$ Non-Class} & \multicolumn{3}{c}{Non-Class $\rightarrow$ Class} & \multicolumn{3}{c}{Random $\rightarrow$ Random}  \\
         \midrule
            &  & 125m & 350m &  1.3b &  125m & 350m &  1.3b & 125m & 350m &  1.3b & 125m & 350m &  1.3b \\
\midrule
     
&  Original  & 44.80 & 41.94 & 42.82 & 38.49 & 42.04 & 42.42 & 46.49 & 44.13 &  44.52 & 42.73 & 42.92 & 44.10 \\
    \midrule
\multirow{13}{*}{\STAB{\rotatebox[origin=c]{90}{\cda}}} & Char. Swap & 44.94 & 42.22 & 42.28 & 39.46 & 40.56 & 42.76 & 47.08 & 44.69 & 44.06 & 43.89 & 42.45 & 42.90 \\
& Char. OCR & 43.72 & 43.70 & 43.66 & 39.31 & 41.02 & \underline{43.73} & 45.91 & 46.02 & 45.11 & 42.95 & 43.02 & 43.83 \\
& Char. Delete & 43.98 & 43.33 & 42.35 & 38.99 & 40.50 & 43.08 & 46.01 & 45.04 & 44.02 & 42.53 & 42.77 & 43.52 \\
& Char. Insert & 45.03 & 43.14 & 41.19 & 39.22 & 40.61 & 42.77 & 46.69 & 44.61 & 42.44 & 43.29 & 42.86 & 42.37 \\
& Char. Subs. & 43.87 & 43.07 & 40.29 & 39.39 & 40.22 & 42.46 & 46.11 & 45.22 & 43.25 & 43.41 & 42.65 & 43.08 \\
& Word Swap & 43.83 & \underline{44.56} & 42.75 & 39.15 & 40.84 & 43.51 & 46.25 & \underline{46.21} & 44.93 & 43.24 & 43.54 & \underline{44.30} \\
& Word Delete & 44.24 & 42.38 & 43.90 & 39.54 & 40.34 & 43.12 & 45.70 & 44.49 & 45.56 & 43.41 & 42.02 & 44.28 \\
& Spell Error & 44.82 & 42.66 & 44.38 & 38.68 & \underline{41.03} & 42.85 & 46.34 & 44.85 & 44.97 & 42.71 & 43.08 & 43.71 \\
& \texttt{GPT2} Insert & 43.37 & 43.36 & 44.13 & \underline{39.79} & 40.92 & 43.69 & 45.07 & 45.93 & 46.11 & 42.65 & 43.07 & 44.44 \\
& \texttt{Word2vec} Insert & \underline{46.47} & 41.80 & 41.86 & 39.67 & 40.80 & 42.66 & \underline{48.36} & 45.13 & 43.78 & \underline{44.18} & 42.95 & 42.86 \\
& \texttt{BERT} Subs. & 44.88 & 44.01 & 44.45 & 39.57 & 40.23 & 42.86 & 46.87 & 45.32 & 46.35 & 43.17 & 42.92 & 43.79 \\
& \texttt{Word2vec} Insert & 44.24 & 43.89 & 42.68 & 38.90 & 40.21 & 42.80 & 46.17 & 45.58 & 44.34 & 43.07 & 42.76 & 44.06 \\
& Back Translation & 44.19 & 44.00 & \underline{46.23} & 39.78 & 40.89 & 42.96 & 45.97 & 45.31 & \underline{47.39} & 43.24 & 43.60 & 43.95 \\
\rowcolor{LightCyan}
\rowcolor{LightCyan}
& \textit{Average}  & 44.42 & 43.23 & 43.08 & 39.34 & 40.62 & 43.01 & 46.34 & 45.26 & 44.79 & 43.21 & 42.89 & 43.62\\
\rowcolor{LightCyan}
& \textit{Best} & 46.47 & 44.56 & 46.23 & 39.79 & 41.03 & 43.73 & 48.36 & 46.21 & 47.39 & 44.18 & 43.47 & 44.30 \\
\midrule
\multirow{13}{*}{\STAB{\rotatebox[origin=c]{90}{\gllm}}} 
& Char. Perturb & 44.72 & 43.56 & 43.30 & 39.39 & 40.61 & 45.54 & 47.02 & 45.32 & 46.25 & 43.68 & 43.27 & 44.06 \\
& Word Insert/Delete & 44.80 & 44.27 & 44.82 & 38.81 & 40.97 & \textbf{44.49} & 46.58 & 45.93 & 46.59 & 42.95 & 43.68 & 44.46 \\
& Word Swap & 45.45 & 43.88 & 45.90 & 39.01 & 40.66 & 43.28 & 47.04 & 45.20 & 46.80 & 43.63 & 42.63 & 45.79 \\
& Word Replace & 45.29 & 45.94 & 46.53 & 39.14 & 41.05 & 43.10 & 47.36 & \underline{47.98} & 47.62 & 43.52 & 44.57 & \underline{45.89} \\
& Grammar Transform & 45.39 & 45.62 & 44.98 & 39.29 & \underline{41.34} & 43.00 & 46.81 & 46.76 & 46.41 & 43.24 & 43.75 & 45.53 \\
& Data Mix & 44.63 & 44.01 & 43.73 & 38.50 & 41.29 & 43.67 & 46.02 & 45.68 & 43.93 & 42.76 & 43.26 & 43.01 \\
& Paragraph Shuffle & 45.78 & 42.03 & 45.80 & 39.29 & 40.70 & 43.28 & 47.70 & 45.02 & \underline{48.02} & \underline{44.43} & 43.25 & 45.41 \\
& Mask Predict & 42.83 & 43.15 & 45.87 & \underline{39.58} & 40.59 & 44.05 & 46.02 & 44.60 & 45.70 & 42.59 & 43.16 & 44.64 \\
& Sentence Reorder & 44.18 & 43.84 & \textbf{48.98} & 39.37 & 40.76 & 43.20 & 46.30 & 46.35 & 48.46 & 43.35 & 43.70 & 45.64 \\
& Contextual Replace & 43.52 & \underline{46.26} & 43.26 & 38.73 & 40.99 & 43.49 & 45.38 & 47.62 & 43.92 & 42.58 & \underline{45.26} & 44.76 \\
& Paraphrase & 45.56 & 45.30 & 42.53 & 39.44 & 40.94 & 41.09 & \underline{48.02} & 47.31 & 43.99 & 43.97 & 44.10 & 42.80 \\
& POS Augment  & 44.97 & 44.48 & 47.51 & 39.12 & 41.14 & 43.57 & 47.06 & 46.07 & 47.25 & 43.49 & 43.21 & 45.75 \\
& Back Translation & \underline{45.79} & 43.38 & 45.30 & 39.18 & 41.12 & 43.80 & 47.02 & 45.58 & 45.10 & 43.36 & 43.20 & 44.41 \\
\rowcolor{LightCyan}
& \textit{Average} & 44.83 & 44.28 & 45.27 & 39.14 & 40.93 & 43.50 & 46.79 & 46.10 & 46.15 & 43.35 & 43.61 & 44.78\\
\rowcolor{LightCyan}
& \textit{Best} & 45.79 & 46.26 & 48.98 & 39.58 & 41.34 & 44.49 & 48.02 & 47.98 & 48.02 & 44.43 & 45.26 & 45.89 \\
    \midrule
 & \textbf{\allm}  & \textbf{51.72} & \textbf{54.98} & {48.80}  & \textbf{40.02} & \textbf{42.80} & {43.80} &  \textbf{50.00} & \textbf{52.75} & \textbf{49.48}  & \textbf{46.64} & \textbf{48.83} & \textbf{48.95} \\
    \bottomrule
    \end{tabular}
    }
    \caption{ The performance of different data augmentation methods.  Char. and Subs. are the abbreviations  of character and substitute, respectively.  \underline{Underlined} indicates best performance under each augmentation method group while \textbf{Bold} indicates the best result of the whole table. In each group, the \colorbox{LightCyan}{last two rows} represent the aggergated result of the whole group of augmentation methods (e.g. average and best).}
    \label{tab:without_adversarial_attack}
    \vspace{-0.35cm}
\end{table*}
\paragraph{Baseline Methods.}
In this study, we compare our novel augmentation pipeline, {\model}, with two different categories of data augmentation  methods as baselines: {\cda} and {\gllm}
For both {\gllm} and {\model}, we employ GPT-3.5 Turbo as the backbone LLM. 
For detailed descriptions of these baseline methods, please see \autoref{sec:aug_instruct}. Specifically:

\begin{itemize}[leftmargin=*,itemsep=0.2pt,topsep=1.pt]
    \item \textbf{{\cda} methods.} This category includes 13 traditional augmentation techniques:
Character-Level: Operations such as random swaps, OCR Errors simulation, deletions, insertions, and substitutions. 
Word-Level: Transformations, including word swaps, deletions, spelling errors, and embedding-based insertions.
Contextual-Level: Utilization of language models for word insertions (e.g., using \texttt{GPT2}~\citep{brown2020language}) and substitutions (e.g., with \texttt{BERT}~\citep{devlin2018bert}),  and back-translation~\citep{fadaee2017data}.

\item \textbf{{\gllm} methods.}  This set comprises 13 manually designed augmentation instructions for LLM, including:
Character-Level: Perturbations similar to those in {\cda}.
Word-Level: Swaps, replacements, and part-of-speech (POS) enhancements.
Sentence-Level: Reordering and data mixing strategies.
Contextual-Level: Predictive masking, contextual substitutions, and back-translation.
\end{itemize}
We also report the average and best performance of {\cda} and {\gllm} for better comparison.
An extensive ablation study of our task-informed selection model, presented in \autoref{sec:ablation}.

\subsection{Main Results}
The analysis of experimental results presented in \autoref{tab:without_adversarial_attack} reveals several findings:
\textbf{Firstly}, there is performance inconsistency among the different instructions from {\gllm}. The impact of augmentation instructions varies across different downstream tasks and models. This highlights the difficulty in creating universally effective data augmentation instruction.
\textbf{Secondly}, {\gllm} is not always better than {\cda}. In controlled comparisons focusing on specific augmentation topics, {\gllm}'s advantages over {\cda} were not clearly evident. For example, in the contexts of ``Back Translation'' and ``Word Swap'', {\cda} outperformed {\gllm} in 5 out of 12 and 7 out of 12 cases, respectively. 
\textbf{Lastly}, the experimental results show the superiority of {\model}. Our proposed model consistently outperformed these baseline methods, highlighting the effectiveness of integrating automatic instruction generation with targeted task-specific instruction selection. This approach not only optimizes performance but also reduces the manual efforts typically required to design effective augmentation strategies, showcasing the potential of our model in enhancing data augmentation practices.

\begin{table}[tbh!]
    \centering
    \small
    \scalebox{0.75}{
    \begin{tabular}{caaa ccc}
    \toprule
        Select. Method & \multicolumn{3}{c}{Class$\rightarrow$Non-Class} & \multicolumn{3}{c}{Non-Class$\rightarrow$Class} \\
        \midrule
        & 125m & 350m &  1.3b & 125m & 350m &  1.3b \\
        \midrule
       {\gllm}$^{+}$   & 39.62 & 41.74 & \textbf{44.38} & 48.02 & 48.51 & 48.07 \\
       \midrule
       \texttt{Random}-Select & 39.34 &    40.31 &  42.68 & 46.15 &               44.34 &             43.98 \\
       \texttt{Empirical}-Select &  39.17 & 41.18             & 43.14  & 47.19             & 47.30              & 44.41    \\
       LLM-Select &  38.77 & 41.06 & 43.07 & 46.81 & 48.02 & 46.42 \\
       \midrule
       {\model}  & \textbf{40.02} & \textbf{42.80} & {43.80} &  \textbf{50.00} & \textbf{52.75} & \textbf{49.48} \\
       \bottomrule
    \end{tabular}
    }
    \caption{Ablation study of {\model}.}
    \label{tab:ablation}
    \vspace{-0.35cm}
\end{table}

\subsection{Ablation Study}
\label{sec:ablation}
We add an ablation study to understand the impact of two key components in our framework: augmentation instruction self-generation and the task-informed instruction selection.
\textbf{Firstly}, we  
% \kz{ not good game}, 
train a task-informed instruction selection model $S$ on manually-crafted instructions from {\gllm} and named it {\gllm}$^+$ to understand the contribution of  the contribution of LLM self-generated augmentation instructions. 
\textbf{Secondly}, we test the efficacy of our selection model by comparing three alternative selection strategies: (1) Random-select, which randomly select instruction from the pool of augmentation methods for each task;
(2) Empirical-select, which selects the prompt that yielded the highest average performance across training tasks, under the assumption that successful prompts on training tasks will generalize well to test tasks; and (3) LLM-Select, which prompts the LLM to chooses the most suitable instruction from candidates based on its internal decision-making processes.
The Results in \autoref{tab:ablation} show that {\model} consistently outperforms these alternative methods,  indicating the benefits of instruction self-generation and task-informed selection in enhancing model performance.

\subsection{Hyperparameter Analysis}
\begin{figure}
    \centering
    \begin{subfigure}[b]{.48\linewidth}
            \includegraphics[width=\linewidth]{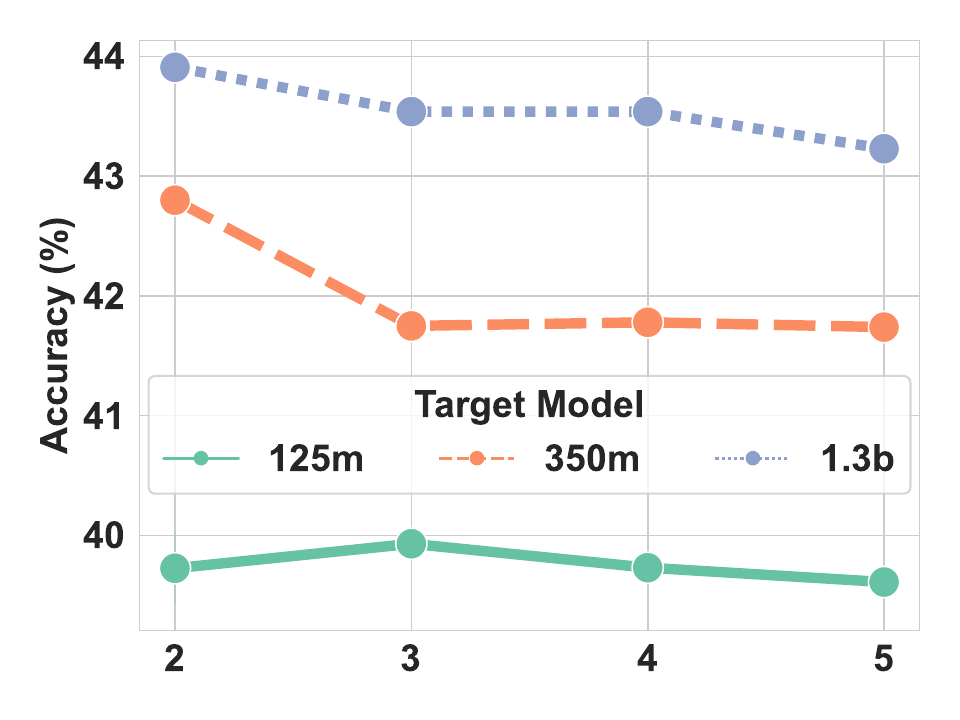}
    \caption{Class$\rightarrow$Non-Class}
    \end{subfigure}
    \begin{subfigure}[b]{.48\linewidth}
            \includegraphics[width=\linewidth]{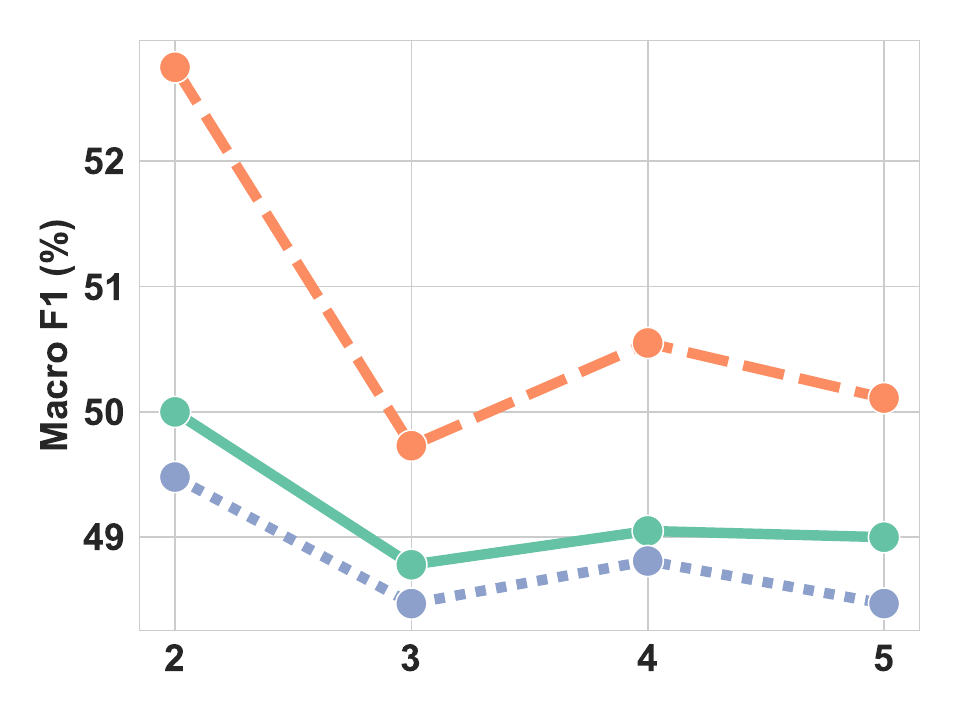}
    \caption{Non-Class$\rightarrow$Class}
    \end{subfigure}
    \caption{Hyperparameter analysis of $n$, which dictates the number of augmentation instructions sampled during the training of the selection model.  }
     \begin{subfigure}[b]{.48\linewidth}
            \includegraphics[width=\linewidth]{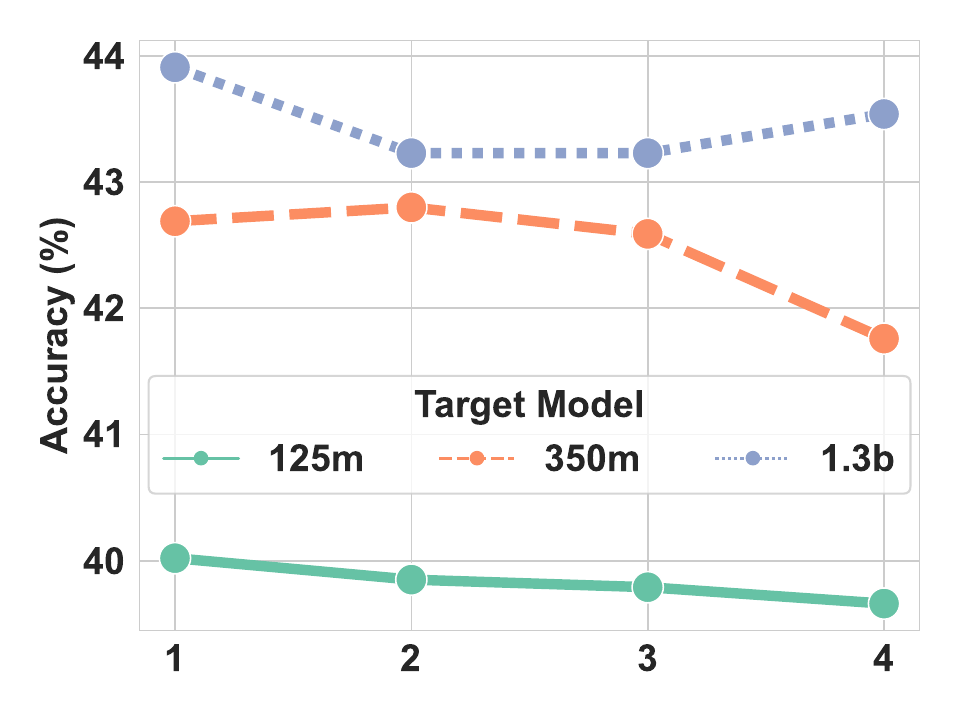}
    \caption{Class$\rightarrow$Non-Class}
    \end{subfigure}
    \begin{subfigure}[b]{.48\linewidth}
            \includegraphics[width=\linewidth]{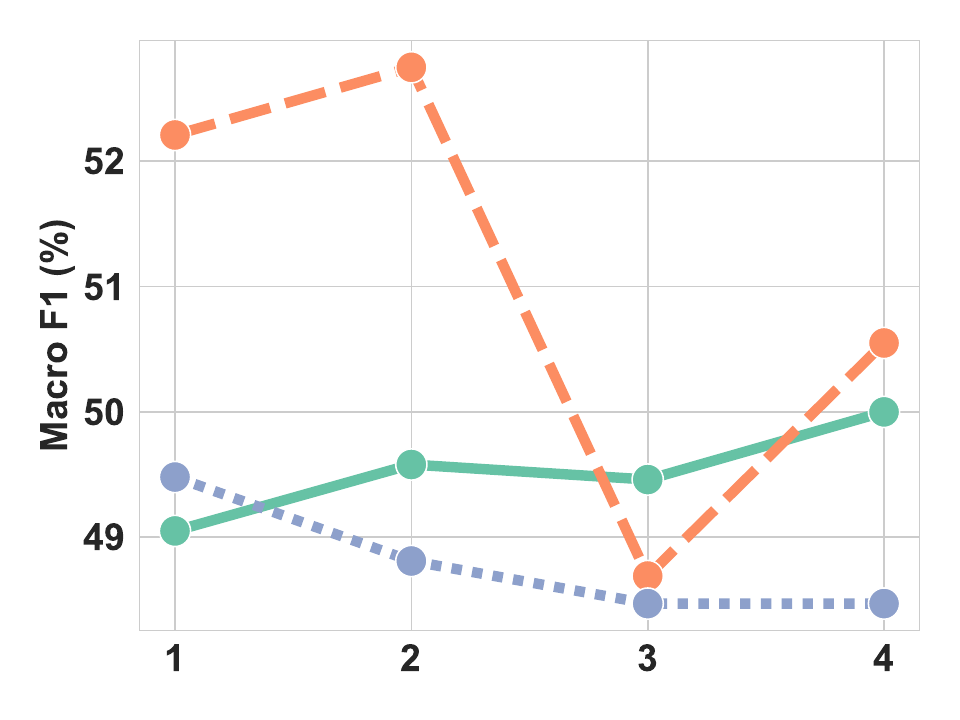}
    \caption{Non-Class$\rightarrow$Class}
    \end{subfigure}
    \caption{Analysis of the hyperparameter $m$, which determines the number of examples randomly sampled to represent a task.}    
    \label{fig:hp_analysis}
    \vspace{-0.5cm}
\end{figure}

Here, we closely examined the impact of two critical hyperparameters on the training of our task-informed instruction model: $n$ and $m$.  
% \kz{why you introduce n first m later}
The hyperparameter $n$ specifies the number of augmentation instructions to be sampled for optimizing \autoref{eq:optimize_s}. It should be noticed that, we only vary $n$ at training time, while at inference, we will calculate the score for all the generated instructions and choose the one with the largest score. 
On the other hand, $m$ determines the number of examples from the task dataset that are used to represent the task, influencing the model's performance during both the optimization and inference phases. 
Our analysis, depicted in \autoref{fig:hp_analysis}, highlights several key findings:
\textit{(1) Optimal Number of Instructions:} We found that setting $n=2$ leads to the best performance, outperforming other configurations. This suggests that a pairwise comparison, as formulated in \autoref{eq:optimize_s}, is most effective for our model's learning process.
\textit{(2) Representative Examples:} Interestingly, a smaller number of examples ($m$) appear to better capture the essence of the tasks. This observation indicates that a larger set of examples could introduce noise, potentially detracting from the model's ability to accurately represent tasks for instruction selection.
\label{sec:hp_analysis}

\subsection{In-Depth Analysis of the Task-Informed Instruction Selection Model}
\label{sec:analysis}
In this section, we provide a detailed analysis of the performance and generalization capabilities of our instruction selection model $S$, focusing on its generalizability to unknown augmentation instructions, unknown target models, and the specific case studies of the augmentation instructions it selects.

\paragraph{Generalization to Unknown Augmentation Instructions.}
In this analysis, we delve into the selection model's adaptability to unknown augmentation instructions by simulating a dynamic environment where new instructions are generated asynchronously by the LLMs. 
This scenario mirrors practical applications where the augmentation instruction set can expand without necessitating retraining of the selection model. 
To test this, we constrained the training phase of the selection model to a limited subset of self-generated augmentation instructions (30\% of all generated by the LLMs), utilizing the whole generated instructions for evaluation at inference time. 

\begin{figure}
    \centering
    \begin{subfigure}[b]{0.48\linewidth}
    \includegraphics[width=\linewidth]{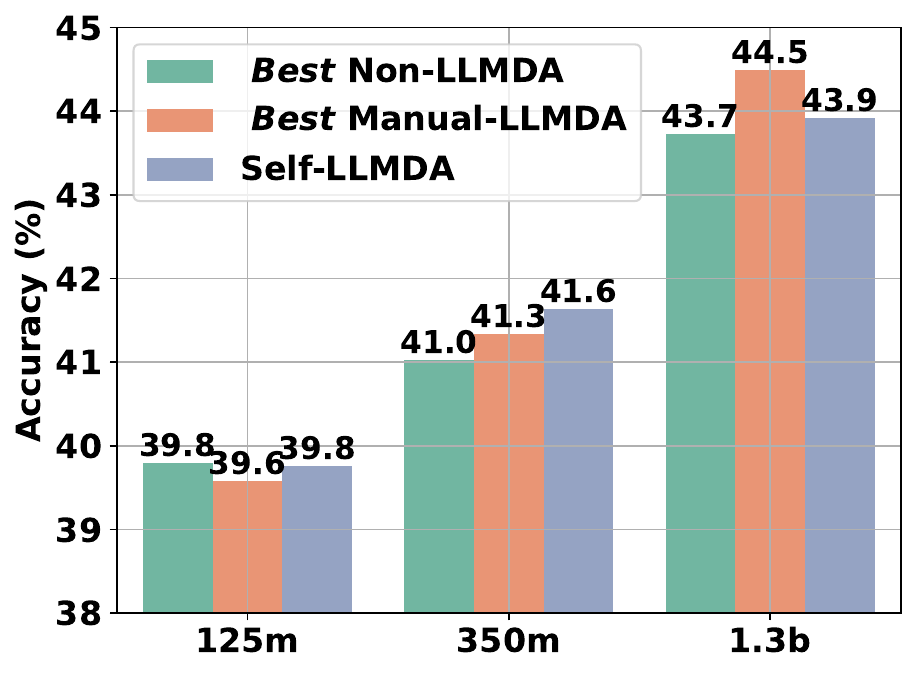}    
    \caption{Class $\rightarrow$ Non-Class.}
    \end{subfigure}
    \begin{subfigure}[b]{0.48\linewidth}
    \includegraphics[width=\linewidth]{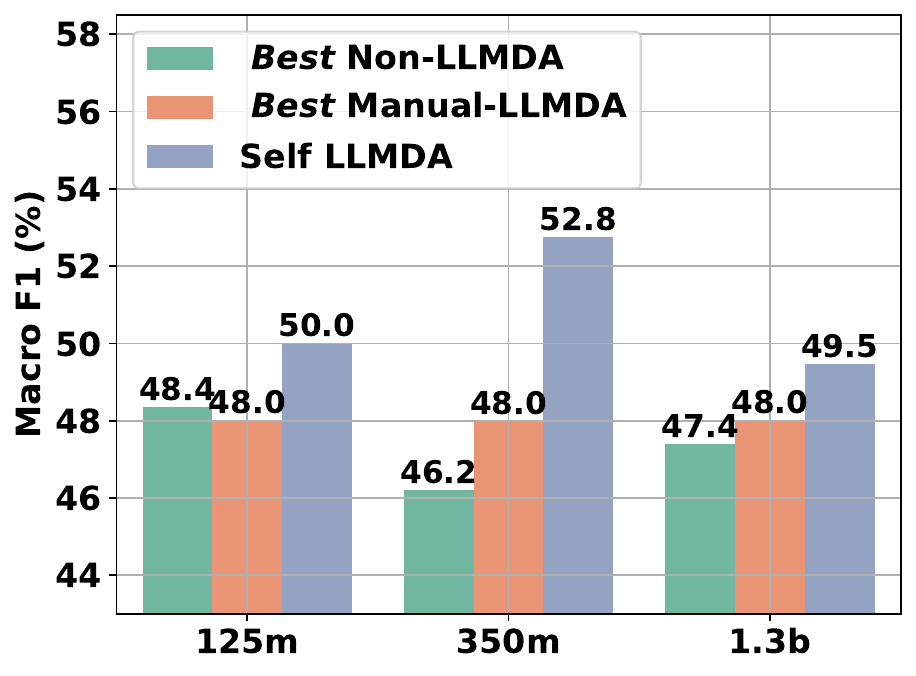}
    \caption{Non-Class $\rightarrow$ Class.
    }
    \end{subfigure}
    
    \caption{Result of generalization to unknown augmentation instruction selection. }
    \label{fig:unknown-prompt}
    \vspace{-0.35cm}
\end{figure}

As the results shown in \autoref{fig:unknown-prompt}, we can observe a performance improvement of our selection model over the best performance of {\cda} and {\gllm}. This indicates the robustness of our selection model in adapting to incremental augmentation instructions, effectively selecting suitable instructions even when faced with previously unknown instructions.  
These observations highlight the efficacy of our selection model in a dynamic augmentation scenario. 
 
\paragraph{Generalization to Unknown Target Models.}
Our study extended to evaluate the adaptability of our task-informed selection model across diverse target models. 
By applying the selection model, initially trained on the task performance of a specific target model, to different models. 
The results of these experiments are presented in \autoref{tab:model-cross-result}.
Our findings show that the augmentation instructions selected by our model remain effective even when applied to different target models.  Notably, in most scenarios, our model {\model}, when transferred to alternate target models, outperformed the best results obtained using {\cda} and {\gllm}. 
This indicates that the underlying pattern determining instruction effectiveness via our instruction selection model is transferable.

\begin{table}[tbh!]
    \centering
    \small
    \scalebox{.75}{
    \begin{tabular}{caaaccc}
    \toprule
        \multirow{2}{*}{Train.}  &  \multicolumn{3}{c}{Class $\rightarrow$ Non-Class} & \multicolumn{3}{c}{Non-Class $\rightarrow$ Class}\\
        & 125m & 350m & 1.3b & 125m & 350m & 1.3b\\
         \midrule
         \rowcolor{LightCyan}
         Best {\cda}  & 39.79 & 41.03 & 43.73 & 48.36 & 46.21 & 47.39 \\
         \rowcolor{LightCyan}
          Best {\gllm}& 39.58 & 41.34 & \textbf{44.49} & 48.02 & 47.98 & 48.02 \\
         \midrule
         125m &  \textbf{40.02} & 41.97 & 43.56 & \textbf{50.00} & \textbf{54.12} & \textbf{49.83} \\
         350m  & 39.96 & \textbf{42.80} &  43.66  & 49.96& 52.75  &48.82\\
         1.3b &  39.85 & 42.42 & {43.80} & 49.04 & 51.22& 49.48  \\
    \bottomrule
    \end{tabular}
    }
    \caption{
        Transferability of the Task-Informed Selection Model. Our selection model, initially trained on a specific target model (indicated by each row in the second group), when applied to alternate target models (represented in each column). 
    }
    \label{tab:model-cross-result}
    \vspace{-0.5cm}
\end{table}

\paragraph{Analysis of Selected Instructions.}
\begin{figure}
    \centering
    \begin{subfigure}[b]{0.49\linewidth}
    \includegraphics[width=\linewidth]{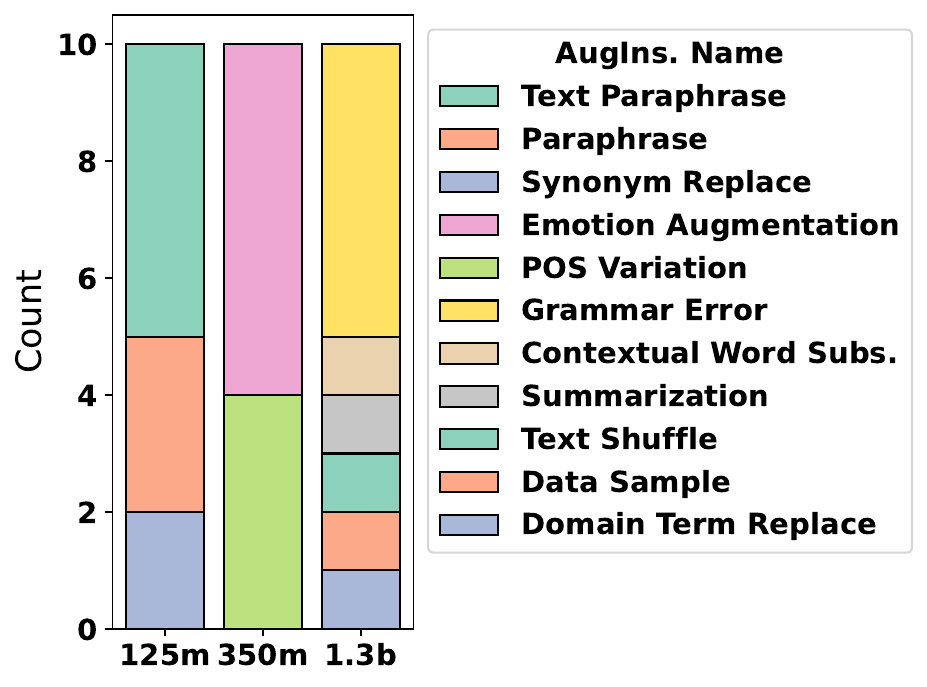}    
    \caption{Class$\rightarrow$Non-Class.}
    \end{subfigure}
        \begin{subfigure}[b]{0.49\linewidth}
    \includegraphics[width=\linewidth]{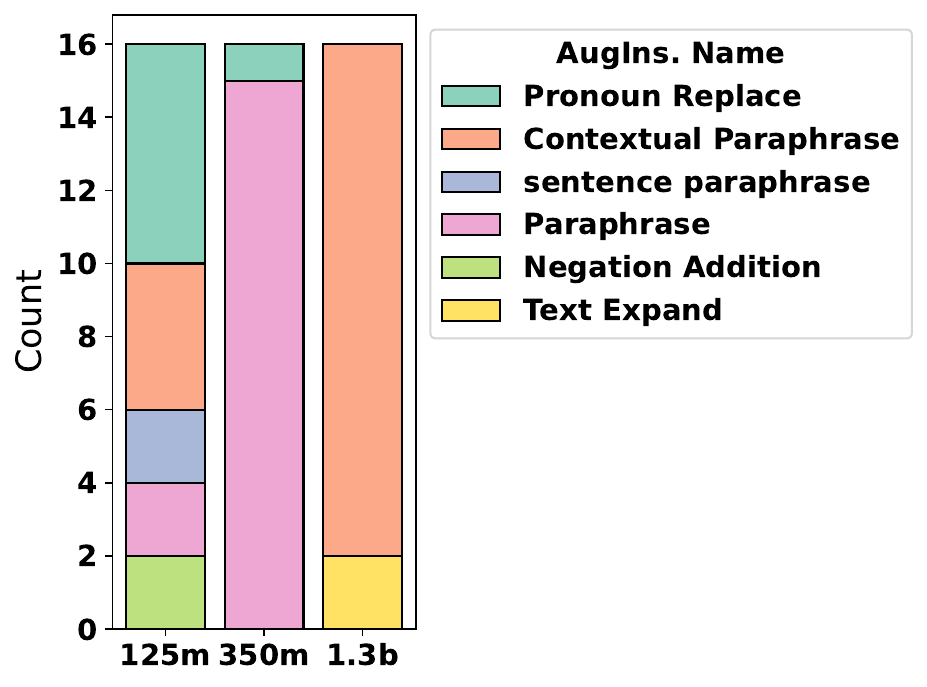}    
    \caption{Non-Class$\rightarrow$Class.}
    \end{subfigure}
    \caption{Selected augmentation instructions from task-informed augmentation selection model. }
    \label{fig:ins-dis}
    \vspace{-0.4cm}
\end{figure}
We conducted a detailed analysis of the augmentation instructions chosen by our selection model, and the findings visualized in \autoref{fig:ins-dis}. The key insights from this analysis are as follows: \textit{(1) Diversity of Selected Instructions:} The distribution of selected instructions showcases a wide variety in the types of augmentations chosen by the model, with 3, 2, and 6 unique data augmentation instructions identified for the 125m, 350m, and 1.3b models under Class$\rightarrow$Non-Class, respectively. This demonstrates the model's ability to adapt and select from a broad spectrum of augmentation strategies to meet the specific requirements of different tasks.
\textit{(2) Variability across Models:} The selection patterns exhibit notable differences when the model is applied to various target models. This variability indicates preference differences across different target models. 
\textit{(3) Preference for Paraphrase-Based Instructions:} A significant portion of the selected instructions fall into the category of paraphrase-based augmentations, such as ``Text Paraphrase'', ``Paraphrase'', ``Contextual Paraphrase'', and ``Sentence Paraphrase''. This preference not only highlights the effectiveness and general applicability of paraphrase-based augmentations but also illustrates our task-informed selection model's nuanced capability to discern and recommend the most suitable paraphrase variation for a given task.

\section{Conclusion}
In this work, we introduced {\model}, a novel framework that leverages the capabilities of LLMs for textual data augmentation. 
Our approach addresses the challenges associated with traditional data augmentation methods and the limitations of manual instruction generation in LLM-based augmentation.  {\model} automates the generation and selection of augmentation instructions, thereby significantly enhancing the quality and applicability of augmented data across diverse downstream tasks.
Tested across 26 diverse few-shot learning tasks, {\model} consistently outperforms both {\cda} and {\gllm} methods, showcasing its effectiveness and applicability.

\clearpage

\section{Limitations}
This study acknowledges several constraints that delineate the scope of our current work and outline directions for future research:

\begin{itemize}
    \item \textbf{Evaluation on a Limited Range of LLMs:} Our experiments were conducted primarily with GPT 3.5 Turbo due to the high costs associated with using OpenAI models. While promising results in \autoref{tab:without_adversarial_attack} suggest that our proposed {\model} method could potentially perform even better on more advanced models like GPT 4 Turbo, comprehensive testing was not feasible. Similarly, the computational demands of evaluating open-source LLMs such as LLAMA-70b-chat~\citep{touvron2023llama}, coupled with the extensive number of tasks in our study, exceeded our resources. Despite these limitations, we are optimistic that {\model} would exhibit enhanced performance across a broader spectrum of LLMs.

    \item \textbf{Meta-Prompting Exploration:} Within the {\model} framework, we employed one meta-prompt to guide the LLM in generating diverse and relevant augmentation instructions. However, our exploration of meta-prompting techniques was limited. We acknowledge that more sophisticated prompt engineering could further refine the quality and effectiveness of generated instructions. Investigating more advanced meta-prompting strategies remains an area for future exploration.

    \item \textbf{Analysis of Ensemble Augmentation Methods:} Our research did not investigate the potential benefits of combining multiple sets of augmented data (e.g., $\mathcal{D} \cup \mathcal{D}'_1 \cup \mathcal{D}'_2$). Such ensemble approaches introduce additional complexities, such as determining the optimal number of augmentation instructions to include. While we hypothesize that ensemble augmentation could improve model performance, this aspect falls outside the current study's scope and is earmarked for subsequent investigation.
\end{itemize}

\bibliography{custom}
\clearpage
\appendix

\section{Detailed Experiment Settings}
\label{sec:detail_exp_setting}
\paragraph{Generation Configuration.}
We utilize gpt-3.5-turbo as our backbone LLM for augmentation instruction generation and data augmentation. We set the temperature for both of them as  0.7. 
For the instruction generation, we follow the generation hyper-parameter setting from \citet{wang2022self}.
For data augmentation, we utilize the default generation hyper-parameter from Chat Completion. The whole experiment including generating augmentation instructions and generating augmentation data costs us \$82 USD in total, according to OpenAI's pricing (	Input \$0.0005 / 1K tokens	and output \$0.0015 / 1K tokens). However, the total experiment cost around \$200 USD for debugging and exploration.    

\paragraph{Meta Prompts for Data Augmentation}
As shown in step \circled{3} in \autoref{fig:model-pipe}, we also need a meta prompt to encourage {\model} to augment high quality data. The main reason for this meta-prompt setting is because in some augmentation instructions they will discuss some external tools like word-embedding, other language models, if we did not provide the meta-prompt, the LLM will reject the generation of augmented data. The design of meta prompt is as follows:
\begin{tcolorbox}[boxsep=0mm,left=2.5mm,right=2.5mm,colframe=black!55,colback=black!5]
\textit{Please do the following data augmentation steps to the text delimited by triple backticks. If you need any external resources or data, you can just simulate the environment by yourself and finish that step based on your own knowledge since you are the best language model in word. \textbf{Augmentation Instructions:} $\mathbf{I}_j$, \textbf{Input Data:} $\mathbf{x}_i$
$\{\mathbf{I}_{\text{seed}}\}$}
\end{tcolorbox}

\paragraph{Task-informed Instruction Selection.}
The instruction ranking model is initialized with FLAN-T5-Large\citep{radford2019language} and is trained using Adafactor~\citep{shazeer2018adafactor} with learning rate 5e-5 and dropout 0.1. We train the selection model for 100 epochs and set the early stop with patience 20 epochs. We employ the validation set from training tasks to select the best checkpoint. The search space for different hyperparameter analysis are as follows:
\begin{table}[h!]
    \centering
    \small
    \scalebox{0.8}{
    \begin{tabular}{c|H|c}
        \toprule
         Symbol& Description & Search Space \\
         \midrule
         $n$ & Number of sampled augmentation instruction & \{2, 3, 4, 5\}\\
         \midrule
              $m$ & Number of sampled examples from task dataset & \{1, 2, 3, 4\} \\
          \midrule
         - & batch size & \{4, 8, 16\} \\
         \midrule
         - & epochs & 100 \\
         \bottomrule
    \end{tabular}
    }
    \caption{The search space of augmentation selection.}
    \label{tab:selection-hp}
\end{table}

\paragraph{Target Model Finetuning.}
We use \texttt{OPT}~\citep{zhang2022opt} from 125m, 350m, 1.3b different sizes. For all of them we use \texttt{AdamW}\citep{loshchilov2019decoupled} as our optimizer with learning rate 5e-5 with 10 training epochs. Due to the constraint of GPU memory, for 125m and 350m we set the batch size as 8, while 1.3b we set the batch size as 2. All these experiments is tested on one NVIDIA A100 A100-40G GPU cards.

\section{Analysis of Self-Generated Instructions }
In our analysis, we delve into the characteristics and diversity of the self-generated augmentation prompts created by {\gllm}. 
\paragraph{Statistical Information.} To facilitate a structured examination, we categorize these prompts based on the textual data augmentation taxonomy outlined by \citet{bayer2022survey}.  
The distribution and basic statistics of these various augmentation methods are detailed in \autoref{tab:statistics}.
\begin{table}[h!]
    \centering
    \small
    \scalebox{0.75}{
    \begin{tabular}{p{0.5\linewidth}cc}
    \toprule
    Augmentation Type & Count & AVG  Length \\
    \midrule
    {\gllm} &  13 & 28.15\\
    \; - character level & 1 & 32.00\\
        \; -  word level & 3 & 20.67 \\
        \; -  phrase level & 4  & 31.75  \\
        \; -  document level &  5 & 29.00 \\
    \midrule
    {\gllm} & 51 & 25.58 \\
    \; - misc. &  1 & 22.00\\
    \; - character level &  2 & 24.50 \\
    \; - word level & 10 & 22.80 \\
    \; - phrase level & 18 & 26.72 \\
    \; - document level & 20 & 26.25  \\
    \bottomrule
    \end{tabular}
    }
    \caption{Statistics of augmentation prompts.}
    \label{tab:statistics}
\end{table}

\begin{figure}
    \centering
    \begin{subfigure}[t]{0.48\linewidth}
        \centering
        \includegraphics[width=\linewidth]{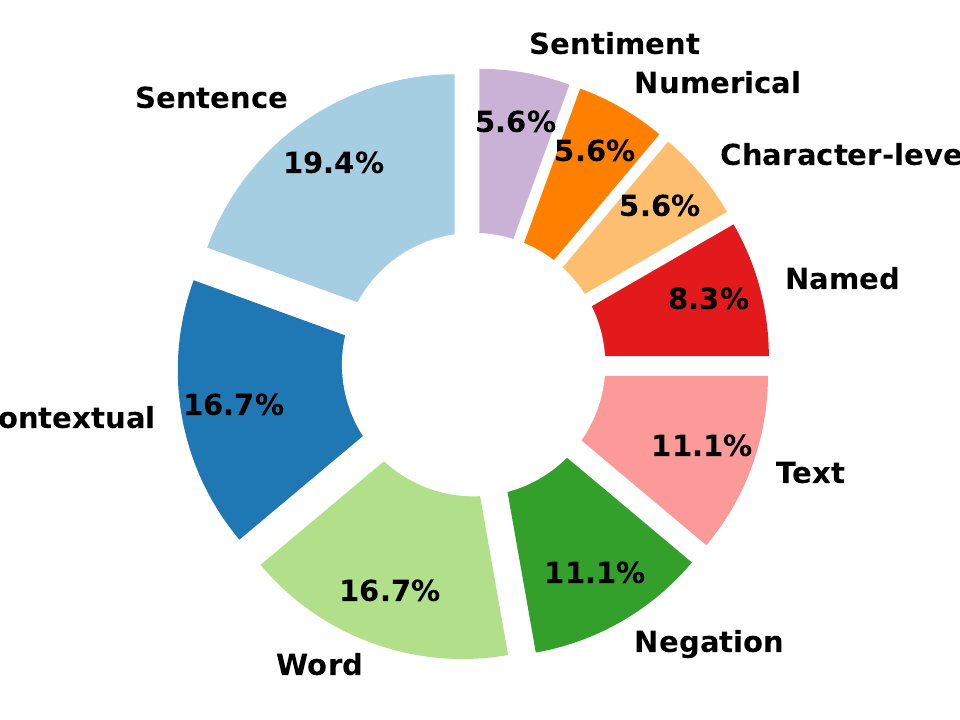}
        \caption{First word.}
    \end{subfigure}  
    \begin{subfigure}[t]{0.48\linewidth}
        \centering
        \includegraphics[width=\linewidth]{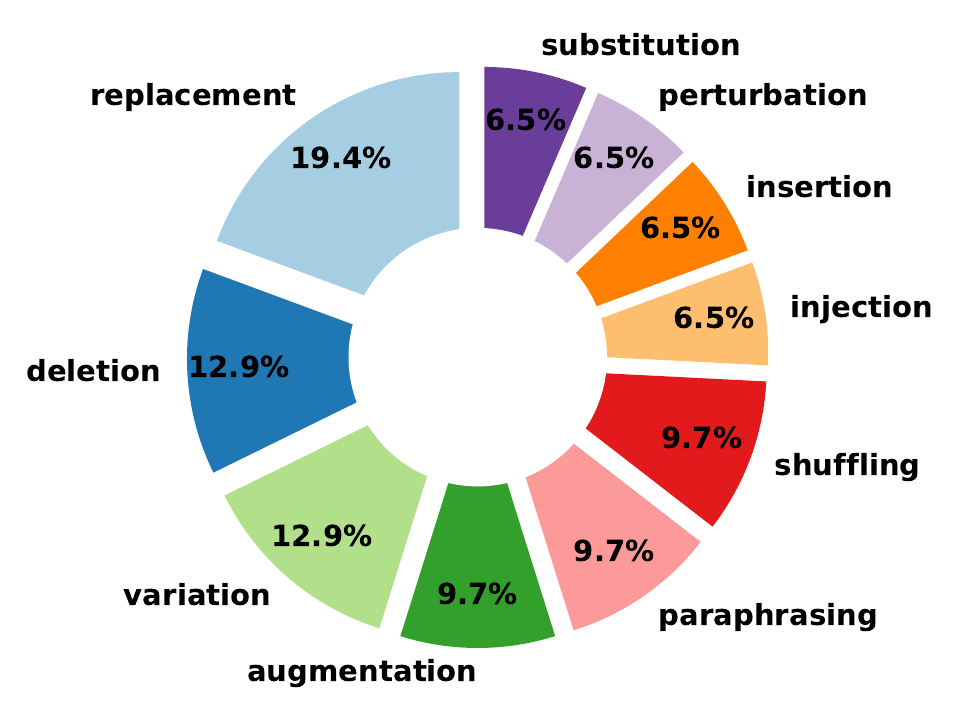}
        \caption{Last word.}
    \end{subfigure}  
    \caption{The first words and last words from the Chat-Self. We filter these words by appearing more than once.  }
    \label{fig:first_last_word_distribution}
\end{figure}

\paragraph{Naming Conventions.}
A notable aspect of our analysis involves examining the naming conventions of the augmentation methods. Recognizing that the method names often provide a high-level summary of the augmentation approach (e.g., <method name>), we further explore the linguistic patterns within these names. Specifically, we conduct an analysis focusing on the first and last words of each method name. This approach allows us to gain insights into the thematic and functional aspects of the augmentation methods. The distribution of these first and last words in method names is visually represented in \autoref{fig:first_last_word_distribution}. This visual representation aids in understanding the range and focus of the augmentation techniques generated by {\gllm}. By analyzing these key linguistic elements, we aim to shed light on the creative breadth and thematic focus of the self-generated augmentation instructions.

\begin{figure}[h!]
    \centering
    \includegraphics[width=0.8\linewidth]{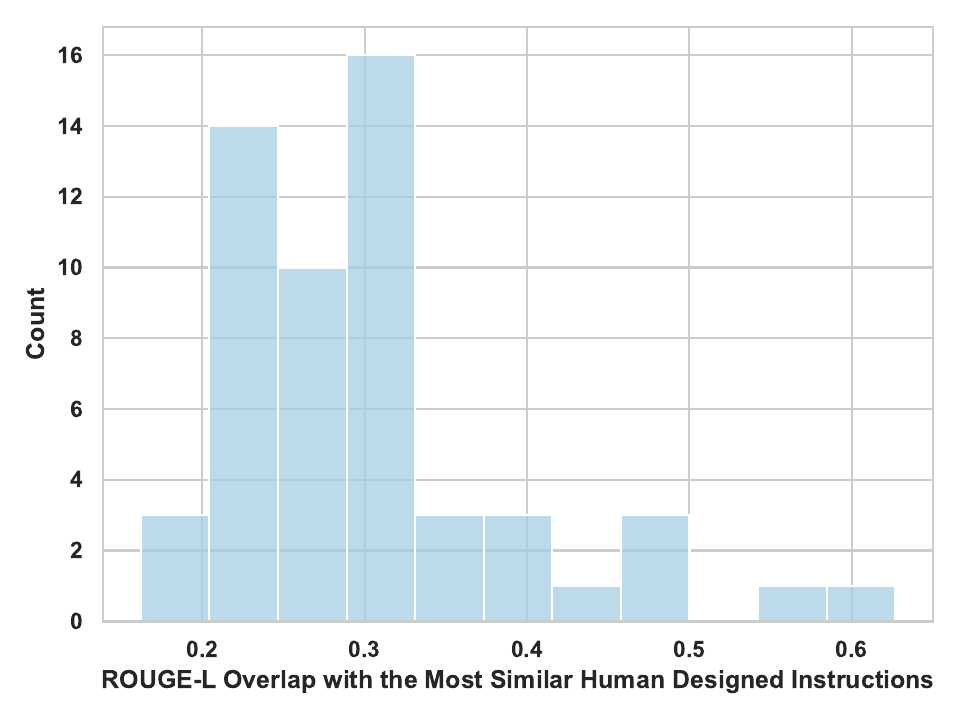}
    \caption{Distribution of the ROUGE-L scores between generated instructions and their most similar human-designed instructions.}
    \label{fig:rouge_l}
\end{figure}

\section{Analysis of Generated Data Across Different Augmentation Methods}
\label{sec:appx_analysis}
In this analysis, we aim to discern the differences among the original dataset, non-LLM augmented data, data augmented via human-designed instructions, and data augmented using {\gllm} generated instructions. 
Our focus is on the surface-level characteristics of the augmented content, and we consolidate data across all tasks for a comprehensive view. Key observations from the analysis, as detailed in \autoref{tab:detail_analysis}, include the following:

\paragraph{Length of Content.} Data augmented by LLM-based methods, on average, exhibits longer content compared to both the original and traditionally augmented datasets. This increase in length could offer a broader spectrum of training examples, potentially aiding in better generalization of target models. However, it also introduces a challenge of dataset inconsistency and the risk of adding unwanted variations.

\paragraph{Perplexity Scores.} Interestingly, LLM-augmented content achieves lower perplexity scores (as measured on GPT2-small) compared to traditional augmentation methods. This suggests that the target model like GPT2-small has a better grasp of content augmented by LLMs. A possible explanation for the higher perplexity scores observed in non-LLM text augmentations is that the character and word-level changes might introduce new, irrelevant tokens into the text, thereby increasing complexity.

\paragraph{Closeness to Original Examples.} Compared to non-LLM augmentation methods, LLM-based augmentations tend to produce content that is more closely related or less diverse relative to the original examples. This observation points to a potential trade-off between relevance and diversity in the augmented content generated by LLMs.

\begin{table}[tbh!]
    \centering
    \small
    \scalebox{0.65}{
    \begin{tabular}{Lcccccc}
    \toprule
        Metrics & Method & Mean & Std. & 25\% & 50\% & 75\% \\
    \midrule
    \multirow{4}{*}{Sentiment} & Ori. & 0.05 & 0.24 & 0.00 & 0.00 & 0.14 \\
    & Traditional & 0.05 & 0.23 & 0.00 & 0.00 & 0.12 \\
         & Human & 0.06 & 0.23 & 0.00 & 0.00 & 0.15 \\
         & ChatGPT & 0.06 & 0.23 & 0.00 & 0.00 & 0.16 \\
    \midrule 
\multirow{2}{*}{Grammar}&     Ori. & 1.50 & 2.34 & 0.00 & 1.00 & 2.00 \\
& Traditional & 4.92 & 5.02 & 2.00 & 4.00 & 7.00 \\
\multirow{2}{*}{Error}& Human & 7.78 & 47.76 & 0.00 & 1.00 & 4.00 \\
& ChatGPT & 4.19 & 27.51 & 0.00 & 1.00 & 3.00 \\
\midrule
\multirow{4}{*}{Words} & Ori. & 29.06 & 38.84 & 11.00 & 20.00 & 34.00 \\
& {\cda} & 30.01 & 39.01 & 11.00 & 21.00 & 36.00 \\
& {\gllm} & 95.34 & 285.95 & 14.00 & 30.00 & 71.00 \\
& {\allm} & 66.04 & 182.91 & 14.00 & 27.00 & 57.00 \\
\midrule
{Perplexity} & Ori. & 530.13 & 5867.93 & 46.42 & 93.78 & 253.85 \\
On & {\cda}& 1354.91 & 5912.87 & 231.22 & 595.03 & 1305.86 \\
GPT2-& {\gllm} & 614.80 & 12284.76 & 17.72 & 58.94 & 184.42 \\
small& {\allm} & 491.83 & 9715.57 & 21.28 & 59.85 & 173.28  \\
\midrule
{Distance} & {\cda} & 0.26 & 0.15 & 0.12 & 0.25 & 0.39 \\
to & {\gllm} & 0.17 & 0.04 & 0.15 & 0.17 & 0.18 \\
Original & {\allm} & 0.15 & 0.05 & 0.12 & 0.15 & 0.18 \\
\bottomrule
    \end{tabular}
    }
    \caption{Characteristics of augmented data. }
    \label{tab:detail_analysis}
\end{table}

\begin{table}[tbh!]
    \centering
    \small
    \scalebox{0.9}{
    \begin{tabular}{cccc}
    \toprule
        Aug. Type & small & medium & large  \\
        \midrule
        Trad.  & 42.37/44.18 &       46.92/49.72 &        44.81/47.33\\
        {Human.} & \textbf{42.43}/44.52 &       47.25/49.52 &        \textbf{48.44}/\textbf{53.32} \\
    {LLM.} & 42.03/\textbf{48.47} &       \textbf{47.80}/\textbf{51.09} &        45.88/52.17 \\
     \bottomrule
    \end{tabular}
    }
    \caption{Main results, using target models from \texttt{GPT2} family. Two numbers indicate the single best augmentation method across tasks and the task specific best augmentation method. Bold indicates the best average result except results.}
    \label{tab:my_label}
\end{table}

\paragraph{Augmentation Instruction Pitfalls Across Tasks}
The effectiveness of augmentation instructions can vary depending on the specific characteristics of the tasks at hand~\citep{ribeiro-etal-2020-beyond, wei2019eda}. To illustrate this, we present a case study focusing on the augmentation instruction \textit{Pronoun replacement: replace pronouns in the text with their corresponding nouns or vice versa, maintaining the semantic meaning of the sentence.}
For the sake of brevity, we will use the abbreviation PR to refer to \textit{pronoun replacement.}
We consider two categories of tasks: text entailment (TE) and question answering (QA). As shown in \autoref{tab:case_study}, the results indicate that PR yields suboptimal performance on TE tasks, while it achieves good performance on QA tasks.
This discrepancy can be attributed to the inherent characteristics of these tasks. 
TE tasks heavily rely on capturing the overall semantic meaning and logical relationships within the text, which may not always be preserved when applying pronoun replacement~\citep{gao2022simcse}. 
In contrast, QA tasks aim to locate and provide specific information relevant to the given question~\citep{joshi2020spanbert}. 
By replacing pronouns with their corresponding nouns, the model can more easily identify the relevant entities and establish a clearer connection between the question and the answer, ultimately benefiting the QA task performance.
\begin{table}[tbh!]
    \centering
    \small
    \begin{tabular}{c|cc}
    \toprule
     Type &  Task & Rank (/51) \\
     \midrule
      \multirow{4}{*}{TE}  & glue-mrpc & 48 \\
        & glue-rte & 44 \\
        & glue-wnli & 33 \\
       & medical\_questions\_pairs & 48 \\
       \midrule
      \multirow{5}{*}{QA} & qasc & 1 \\
      & openbookqa & 27 \\
      & commonsense\_qa & 4 \\
      & quartz-no\_knowledge & 34 \\
      & quartz-with\_knowledge & 9 \\
        \bottomrule
    \end{tabular}
    \caption{Performance comparison of the pronoun replacement (PR) augmentation instruction on text entailment (TE) and question answering (QA) tasks.}
    \label{tab:case_study}
\end{table}

\section{Dataset Collection}
\label{sec:dataset_details}
In \autoref{tab:tasks_all}, we list all 26 tasks and how we splitting them into training and testing for evaluating the model generalization to unknown downstream tasks. Each task will have 16 training and validation examples but with full test examples. We utilize the code  from \texttt{CrossFit}\citep{ye-etal-2021-crossfit} to extract and split the training, validation and testing for each task. 

\begin{table*}[tbh!]
    \centering
    \small
    \scalebox{0.8}{
    \begin{tabular}{cc|cccc}
    \toprule
    Category & Train/Test & \multicolumn{4}{c}{Task Names}\\
    \midrule
    \multirow{4}{*}{Class $\rightarrow$ Class}   & Train & financial\_phrasebank &  ethos-religion & glue-wnli & glue-mrpc \\
    \cline{2-6} 
        & \multirow{3}{*}{Test} & tweet\_eval-stance\_feminist & climate\_fever & poem\_sentiment & tweet\_eval-hate \\
        & & ethos-race&
                          tweet\_eval-stance\_atheism& sick& glue-rte \\
                          && superglue-cb& ethos-national\_origin&
                          medical\_questions\_pairs& hate\_speech18 \\

    \midrule
   \multirow{7}{*}{Class$\rightarrow$Non-Class} & \multirow{4}{*}{Train} & tweet\_eval-stance\_atheism& superglue-cb& financial\_phrasebank& ethos-religion \\
    && tweet\_eval-stance\_feminist& climate\_fever& glue-mrpc& tweet\_eval-hate\\
    && glue-rte&
ethos-national\_origin& glue-wnli& medical\_questions\_pairs \\
&& sick& poem\_sentiment& hate\_speech18& ethos-race\\
\cline{2-6} 
& \multirow{3}{*}{Test} & quarel&  codah&  ai2\_arc&  openbookqa \\
&& superglue-copa&  qasc&  quartz-no\_knowledge&  dream \\
&& quartz-with\_knowledge&  commonsense\_qa\\
\midrule
\multirow{7}{*}{Non-Class $\rightarrow$ Class} & \multirow{3}{*}{Train} & quarel&  codah&  ai2\_arc&  openbookqa \\
&& superglue-copa&  qasc&  quartz-no\_knowledge&  dream \\
&& quartz-with\_knowledge&  commonsense\_qa\\
\cline{2-6} 
& \multirow{4}{*}{Test} & tweet\_eval-stance\_atheism& superglue-cb& financial\_phrasebank& ethos-religion \\
    && tweet\_eval-stance\_feminist& climate\_fever& glue-mrpc& tweet\_eval-hate\\
    && glue-rte&
ethos-national\_origin& glue-wnli& medical\_questions\_pairs \\
&& sick& poem\_sentiment& hate\_speech18& ethos-race\\
\midrule
\multirow{7}{*}{Random $\rightarrow$ Random}& \multirow{2}{*}{Train} &  quartz-with\_knowledge&  tweet\_eval-stance\_feminist&  codah&   ethos-race \\ 
&& financial\_phrasebank&  ai2\_arc&  superglue-cb\\
\cline{2-6}
& \multirow{5}{*}{Test} & quartz-no\_knowledge& hate\_speech18& medical\_questions\_pairs& 
ethos-religion \\
&& glue-rte& commonsense\_qa& superglue-copa& ethos-national\_origin \\
&& glue-mrpc& poem\_sentiment& quarel& dream \\
&& climate\_fever& tweet\_eval-hate& qasc& glue-wnli \\
&& tweet\_eval-stance\_atheism& openbookqa& sick\\
         \bottomrule
    \end{tabular}
    }
    \caption{All tasks used in this paper. We split them into training and testing sets under different experiment setting. }
    \label{tab:tasks_all}
\end{table*}

\section{Details of Baseline Methods}
\label{sec:aug_instruct}
\subsection{Augmentation Methods of \cda}
All of the implementation of {\cda} are from \citet{ma2019nlpaug}. Here is an elaboration on each of the mentioned {\cda} augmentation methods: 
\paragraph{}{Character-Level Augmentations} Random Swap~\citep{belinkov2018synthetic}: This involves swapping adjacent characters within words to simulate typos that might occur during typing. For example, "example" might become "exmaple". 
OCR Replace: Simulating errors commonly introduced by Optical Character Recognition (OCR) software when digitizing text. Characters that look similar, like 'o' and '0' or 'l' and '1', might be substituted for one another.
Delete: Randomly removing characters from words to mimic typographical errors or omissions.
Insert: Adding extra characters into words at random positions, simulating common typos or spelling errors.
Substitute: Replacing characters in words with other characters, not necessarily similar in appearance, to create variations in the text.
\paragraph{Word-Level Augmentations.} Swap~\citep{wei2019eda}: Changing the positions of two adjacent words in a sentence to add syntactic variability while largely preserving the sentence's meaning.
Delete: Removing words from sentences randomly to simulate information loss and encourage the model to learn from incomplete data.
Spell Error~\citep{coulombe2018text}: Introducing common spelling mistakes into words to mimic human error and increase the model's exposure to varied spellings.
Word2Vector Insert~\citep{morris2020textattack}: Identifying suitable locations in a sentence to insert synonyms or related words based on word embeddings (like word2vec representations), enhancing semantic diversity.
\paragraph{Contextual-Level Augmentations} Insert Word using \texttt{GPT2}~\citep{kumar2021data}: Leveraging a pre-trained model like \texttt{GPT2} to generate contextually relevant words to insert into sentences, increasing the complexity and variability of the sentence structures.
Substitute Word using \texttt{BERT}~\citep{kumar2021data}: Using a model like \texttt{BERT} to identify and replace words with contextually appropriate synonyms or related terms, maintaining the sentence's overall meaning while altering its surface form.
Back-Translation~\citep{fadaee2017data}: Translating a sentence into another language and then back into the original language. This process often introduces syntactic and lexical variations, providing a paraphrased version of the original sentence that retains its semantic content.

\section{Additional Experiment}
We also compare our method with other data augmentation techniques from {\cda} and {\gllm}. The {\cda} includes EDA~\citep{wei2019eda} and AEDA~\citep{karimi-etal-2021-aeda-easier}, while {\gllm} includes GPT3Mix~\citep{yoo2021gpt3mix} and ZeroGen~\citep{ye2022zerogen}. As  shown in \autoref{tab:more_experiment}, our proposed method {\model} significantly outperforms these baseline methods in the Class$\rightarrow$Class and Random$\rightarrow$Random settings.  However, in the Non-Class$\rightarrow$Class setting, {\model} falls behind GPT3Mix. This may indicate suboptimal transferability of {\model} in this specific scenario. It is worth noting that GPT3Mix is designed specifically for classification tasks, whereas {\model} can be applied to a wide range of text-related tasks, demonstrating its versatility and broader applicability.

\begin{table*}[tbh!]
    \centering
    \small
    \begin{tabular}{c aa cc aa cc}
    \toprule
        \multirow{2}{*}{TextualDA} &  \multicolumn{2}{a}{Class$\rightarrow$ Class}& \multicolumn{2}{c}{Class$\rightarrow$ Non-Class} & \multicolumn{2}{a}{Non-Class$\rightarrow$ Class} & \multicolumn{2}{c}{Random$\rightarrow$Random}\\
        &  125m & 350m &  125m & 350m & 125m & 350m & 125m & 350m \\
        \midrule
        EDA & 45.28 & 44.55 & 39.34 & 41.27 & 47.60& 46.97 & 43.89 &  44.16 \\
        AEDA & 43.85 & 42.92 & 39.56 & 41.53 & 42.61  & 43.50  & 42.61 &    43.50 \\
        ZeroGen-LLM & 45.66 &  45.81 & \textbf{40.43} &            40.62 & 47.69 & 48.58 & 44.14 & 44.74 \\
        GPT3Mix & 51.62 & 51.36 & - & - & \textbf{55.81} & \textbf{54.04} & - & - \\
        \midrule 
        {\allm} & \textbf{51.72} & \textbf{54.98}  & {40.02} & \textbf{42.80} &{50.00} & {52.75} & \textbf{46.64} & \textbf{48.83} \\
    \bottomrule
    \end{tabular}
    \caption{Performance comparison with other non-LLM-based and LLM-based textual data augmentations.}
    \label{tab:more_experiment}
\end{table*}

\subsection{Augmentation Instructions of \gllm}
\label{sec:human_ins}
In \autoref{tab:human-designed}, we will represent the manually crafted augmentation instructions. The format of these augmentation instructions is ``<method name>:
<method instruction>''.

\begin{table*}[tbh!]
    \centering
    \small
    \scalebox{0.8}{
    \begin{tabular}{c|B}
    \toprule
        Name &  Augmentation Instruction \\
        \midrule
        Word Replace  &  \textbf{Synonym Replacement:} Replace certain words in the text with their synonyms while keeping the sentence structure intact. This can be done using pre-built synonym databases or word embeddings. \\
        \midrule
        Back Translation & \textbf{Back Translation:} Translate the text into another language using machine translation and then translate it back to the original language. This process introduces variations in the sentence structure and wording. \\
        \midrule
        Paraphrase & \textbf{Paraphrase:} Render the same text in different words without losing the meaning of the text itself. More often than not, a paraphrased text can convey its meaning better than the original words. \\      
        \midrule
        Word Insert/Delete & \textbf{Random Insertion/Deletion:} Randomly insert or delete words in the text to create new variations of the original sentences.\\
        \midrule
        Word Swap & \textbf{Random Swapping:} Randomly swap the positions of words in the text to create new sentence arrangements. \\
        \midrule
        Mask Predict &  \textbf{Masking/Prediction:} Mask certain words in the text and train the model to predict those masked words. This is similar to the concept of masked language modeling used in models like BERT.\\
        \midrule
 Char. Perturb & \textbf{Character-level Perturbation:} Instead of operating at the word level, perform data augmentation at the character level. This can involve randomly replacing, inserting, or deleting characters within the text, leading to novel variations. \\
 \midrule
Sentence Reorder & \textbf{Sentence Reordering:} Randomly reorder the sentences in the text while maintaining the coherence of the overall passage. This can help the model become more robust in understanding different sentence arrangements.\\
\midrule
Contextual Replace & \textbf{Contextual Synonym Replacement:} Instead of blindly replacing words with synonyms, consider the context of the sentence to choose appropriate synonyms. This can be achieved by using contextual word embeddings or language models like ELMo or BERT. \\
\midrule
POS Augment & \textbf{Part-of-Speech Augmentation:} Identify the part-of-speech tags of words in the text and replace words with synonyms that have the same part-of-speech. This ensures that the grammatical structure of the sentence remains intact. \\
\midrule
Grammar Transform & \textbf{Grammar Transformation:} Apply various grammar rules to the text, such as changing active voice to passive voice, transforming affirmative sentences to negative, or converting declarative sentences into questions. \\
\midrule
Data Mix & \textbf{Data Mixing:} Combine two or more texts from different sources to create a new mixed-text data point. This can introduce diversity in the content and writing style. \\
\midrule
 Paragraph Shuffle & \textbf{Paragraph Shuffling:} Shuffle the order of paragraphs in longer texts to create new document structures. This can be particularly useful for tasks that involve document-level understanding. \\
\midrule
         \bottomrule
    \end{tabular}
    }
    \caption{Manually crafted augmentation instructions.}
    \label{tab:human-designed}
\end{table*}

\section{Self Instructions Generation}
The instructions automatically generated by LLM is shown in \autoref{tab:auto_gen0} and \autoref{tab:auto_gen1}.
\begin{table*}[tbh!]
    \centering
    \small
    \scalebox{0.85}{
    \begin{tabular}{c|B}
    \toprule
         &  Augmentation Instruction \\
        \midrule
&  Contextual word replacement: replace certain words in the text with contextually similar words. this can be done by using word embeddings to find words that have similar meanings or are used in similar contexts. \\ \midrule &  Sentiment flipping: change the sentiment of the text by flipping positive sentiments to negative and vice versa. this can help generate diverse data for sentiment analysis tasks. \\ \midrule &  Style transfer: transform the writing style of the text while maintaining its original content. this can involve converting formal language to informal, changing the tone, or adapting the text to a specific genre. \\ \midrule &  Contextual paraphrasing: paraphrase sentences in the text while considering the surrounding context. this ensures that the paraphrased sentences maintain the same meaning within the given context. \\ \midrule &  Domain adaptation: modify the text to make it more suitable for a different domain or topic. this can involve replacing domain-specific terms, adjusting terminology, or incorporating relevant keywords. \\ \midrule &  Text summarization: generate a summary of the text by condensing its main points into a shorter version. this can help create diverse data for summarization tasks and provide alternative perspectives on the original text. \\ \midrule &  Contextual word insertion: insert new words into the text that are contextually relevant and maintain the semantic meaning of the sentence. this can be done by using word embeddings to find words that are commonly used in similar contexts. \\ \midrule &  Text paraphrasing: paraphrase the text by rephrasing sentences while preserving the original meaning. this can be done using techniques such as sentence splitting, word substitution, and sentence rearrangement. \\ \midrule &  Contextual sentence deletion: delete certain sentences from the text while maintaining the coherence and semantic meaning of the remaining sentences. this can help create diverse data for document summarization tasks. \\ \midrule &  Text expansion: expand the text by adding additional details, examples, or explanations to enhance its content. this can be done by incorporating information from external sources or generating new sentences based on the existing text. \\ \midrule &  Emotion augmentation: add emotional expressions or sentiments to the text to convey different emotions. this can help create diverse data for emotion classification or sentiment analysis tasks. \\ \midrule &  Named entity substitution: replace named entities (such as names of people, organizations, or locations) in the text with other similar entities. this can introduce variations while maintaining the context of the sentence. \\ \midrule &  Sentence reordering: rearrange the order of sentences within the text while keeping the semantic coherence intact. this can create new narrative structures and perspectives. \\ \midrule &  Grammatical error injection: introduce grammatical errors into the text by randomly modifying verb tenses, subject-verb agreement, or punctuation. this can simulate noisy real-world data and improve model robustness. \\ \midrule &  Word embedding replacement: replace certain words in the text with their word embeddings. this can introduce variations while preserving the semantic meaning of the sentence. \\ \midrule &  Synonym replacement: replace words in the text with their synonyms. this can create alternative phrasing while maintaining the overall meaning. \\ \midrule &  Contextual word substitution: substitute words in the text with other words that have similar contextual meanings. this ensures that the replacement maintains the intended semantics. \\ \midrule &  Sentence splitting: split longer sentences into shorter ones, or vice versa, to create new sentence structures. this can help the model understand different sentence lengths and improve its generalization ability. \\ \midrule &  Sentence combination: combine multiple shorter sentences into a single longer sentence or vice versa. this can create diverse sentence structures and test the model\'s comprehension of complex sentences. \\ \midrule &  Negation insertion: introduce negations into the text by adding words like "not" or "no" to change the polarity of certain statements. this can help the model understand negative contexts better. \\ \midrule &  Word masking: randomly mask certain words in the text by replacing them with a special token. this forces the model to rely on the surrounding context to understand the meaning of the masked word. \\ \midrule &  Character-level augmentation: modify individual characters within words, such as changing vowels or consonants, to generate diverse textual variations. \\ \midrule &  Sentence shuffling: shuffle the order of sentences within a paragraph or document to create new arrangements and test the model\'s ability to understand different contexts. \\ \midrule &  Paraphrasing: rewrite the text using different phrasing or sentence structures while maintaining the original meaning. \\ \midrule &  Numerical value perturbation: add or subtract small random values to numerical values in the text to create slight variations. \\ \midrule &  Pos tagging variation: randomly change the part-of-speech tags of words in the text while ensuring grammatical correctness. this can introduce different syntactic patterns and word usages. \\ \midrule &  Sentence deletion: remove one or more sentences from the text to create a shorter version while still maintaining coherence and semantic meaning. \\ \midrule &  Sentiment modification: change the sentiment or emotion expressed in the text while keeping the content intact. this can involve altering positive statements to negative ones or vice versa. \\ \midrule &  Negation addition: add negation words (e.g., "not," "never") to the text to create negative versions of the original sentences. \\ \midrule &  Word order variation: randomly change the order of words within phrases or clauses in the text to generate new sentence arrangements. \\ \midrule &  Word sense disambiguation: replace ambiguous words in the text with their different senses to create diverse interpretations of the sentence.\\
         \bottomrule
    \end{tabular}
    }
    \caption{Automatic generated augmentation instructions. }
    \label{tab:auto_gen0}
\end{table*}

\begin{table*}
    \centering
    \small
    \scalebox{0.85}{
    \begin{tabular}{c|B}
    \toprule
         &  Augmentation Instruction \\
        \midrule
      &  Negation transformation: transform positive statements into negative ones by adding negation words or phrases, or vice versa. \\ \midrule &  Contradiction generation: introduce contradictions within the text by modifying certain statements to be opposite to what they originally conveyed. \\ \midrule &  Named entity replacement: identify named entities in the text (e.g., names of people, organizations) and replace them with similar entities to generate new variations. \\ \midrule &  Word order shuffling: randomly shuffle the order of words within a sentence to create new sentence structures. \\ \midrule &  Grammar error injection: introduce grammatical errors such as incorrect verb conjugation, subject-verb agreement, or punctuation mistakes to simulate natural language variation. \\ \midrule &  Text shuffling: shuffle the order of sentences within a paragraph or paragraphs within a document to create new arrangements. this helps diversify the structure and flow of the text. \\ \midrule &  Contextual deletion: remove certain words or phrases from the text while ensuring that the remaining content still conveys the same overall meaning. this tests the model\'s ability to understand and fill in missing information. \\ \midrule &  Sentence paraphrasing: generate paraphrases of the original sentences while preserving their meaning. this can be achieved through techniques like back-translation or paraphrase generation models. \\ \midrule &  Named entity variation: replace named entities (such as names, locations, organizations) in the text with different variations to create diverse instances of the same sentence. \\ \midrule &  Numerical variation: modify numerical values in the text by adding/subtracting a small random value or replacing them with synonyms/alternative representations. \\ \midrule &  Negation augmentation: introduce negations in the text by adding "not" or other negation words to certain phrases or sentences. this helps the model understand negative contexts better. \\ \midrule &  Data sampling: randomly sample subsets of the data to create smaller training sets for faster experimentation and exploration of different data distributions. \\ \midrule &  Backtranslation: translate the text into another language and then translate it back to the original language. this can introduce variations in sentence structure and word choice while preserving the overall meaning. \\ \midrule &  Sentence splitting/merging: split long sentences into shorter ones or merge short sentences into longer ones to create new sentence structures. \\ \midrule &  Pronoun replacement: replace pronouns in the text with their corresponding nouns or vice versa, maintaining the semantic meaning of the sentence. \\ \midrule &  Word deletion: randomly delete certain words from the text to create shorter or more concise sentences. this can simulate scenarios where some information is missing or incomplete. \\ \midrule &  Character-level perturbation: introduce noise at the character level by randomly changing, deleting, or inserting characters in the text. this can help models become more robust to noisy inputs and improve generalization. \\ \midrule &  Domain-specific term replacement: identify domain-specific terms in the text and replace them with synonyms or related terms specific to another domain. this can help models generalize better across different domains. \\ \midrule &  Negation/positive conversion: convert negative statements to positive ones or vice versa to generate different perspectives or sentiments. \\ \midrule &  Part-of-speech tagging: modify the part-of-speech tags of words in the text to create new grammatical arrangements and sentence structures.  \\
     \bottomrule
     \end{tabular}
         }
         \caption{Automatic generated augmentation instructions.}
         \label{tab:auto_gen1}
\end{table*}

\end{document}